\documentclass{article}
\usepackage{graphicx}
\usepackage{bbm}
\usepackage{mathtools}
\usepackage{amsmath}
\PassOptionsToPackage{sort&compress}{natbib} 
\usepackage{color}
\usepackage{wrapfig}

\usepackage{algorithm}
\usepackage{algpseudocode}
\usepackage{algorithmicx}
\usepackage{listings}
\usepackage{amssymb}
\usepackage{bm}
\usepackage{titlesec}
\usepackage{caption}
\DeclareMathOperator*{\argmin}{argmin}

\usepackage{titlesec}
\titlespacing\section{0pt}{3pt plus 4pt minus 2pt}{0pt plus 2pt minus 2pt}
\titlespacing\subsection{0pt}{3pt plus 4pt minus 2pt}{0pt plus 2pt minus 2pt}
\titlespacing\subsubsection{0pt}{3pt plus 4pt minus 2pt}{0pt plus 2pt minus 2pt}

\usepackage{hyperref}

\usepackage{microtype}
\usepackage{subfigure}
\usepackage{booktabs} 

\DeclareMathAlphabet{\mathdutchcal}{U}{dutchcal}{m}{n}

\usepackage{mleftright,xparse}

\DeclarePairedDelimiterX{\infdivx}[2]{\mleft(}{\mright)}{%
  #1\;\delimsize\|\;#2%
}

\newcommand{\sgiven}{\,|\,}
\newcommand{\given}{\,\middle|\,}
\newcommand{\kld}{\vert\vert\,}

\usepackage{authblk}
\usepackage{soul}
\usepackage[dvipsnames]{xcolor}

\setstcolor{blue}

\usepackage[final]{corl_2019} 

\title{Entity Abstraction in Visual Model-Based Reinforcement Learning}

\author{
  \textbf{Rishi Veerapaneni$^{*,1}$}, \textbf{John D. Co-Reyes$^{*,1}$}, \textbf{Michael Chang$^{*,1}$}, \textbf{Michael Janner$^{1}$}
  \\\textbf{Chelsea Finn$^{2}$},  \textbf{Jiajun Wu$^{3}$}, \textbf{Joshua B. Tenenbaum$^{3}$}, \textbf{Sergey Levine$^{1}$}
}

\begin{document}
\let\svthefootnote\thefootnote
\let\thefootnote\relax\footnote{* Equal contribution.}
\let\thefootnote\relax\footnote{$^1$University of California Berkeley. $^2$Stanford University. $^3$Massachusetts Institute of Technology.}
\let\thefootnote\relax\footnote{Code and visualizations are available at \url{https://sites.google.com/view/op3website/}.}
\let\thefootnote\relax\footnote{This article is an extended version of the manuscript published in CoRL 2019.}
\let\thefootnote\svthefootnote
\maketitle


\begin{abstract}
This paper tests the hypothesis that modeling a scene in terms of entities and their local interactions, as opposed to modeling the scene globally, provides a significant benefit in generalizing to physical tasks in a combinatorial space the learner has not encountered before. We present object-centric perception, prediction, and planning (OP3), which to the best of our knowledge is the first fully probabilistic entity-centric dynamic latent variable framework for model-based reinforcement learning that acquires entity representations from raw visual observations without supervision and uses them to predict and plan. OP3 enforces entity-abstraction -- symmetric processing of each entity representation with the same locally-scoped function -- which enables it to scale to model different numbers and configurations of objects from those in training. Our approach to solving the key technical challenge of grounding these entity representations to actual objects in the environment is to frame this variable binding problem as an inference problem, and we develop an interactive inference algorithm that uses temporal continuity and interactive feedback to bind information about object properties to the entity variables. On block-stacking tasks, OP3 generalizes to novel block configurations and more objects than observed during training, outperforming an oracle model that assumes access to object supervision and achieving two to three times better accuracy than a state-of-the-art video prediction model that does not exhibit entity abstraction.
\end{abstract}



\section{Introduction} \label{sec:intro}
A powerful tool for modeling the complexity of the physical world is to frame this complexity as the composition of simpler entities and processes.
For example, the study of classical mechanics in terms of macroscopic objects and a small set of laws governing their motion has enabled not only an explanation of natural phenomena like apples falling from trees but the invention of structures that never before existed in human history, such as skyscrapers.
Paradoxically, the creative \textit{variation} of such physical constructions in human society is due in part to the \textit{uniformity} with which human models of physical laws apply to the literal building blocks that comprise such structures -- the reuse of the same simpler models that apply to primitive entities and their relations in different ways obviates the need, and cost, of designing custom solutions from scratch for each construction instance.

The challenge of scaling the generalization abilities of learning robots follows a similar characteristic to the challenges of modeling physical phenomena: the complexity of the task space may scale combinatorially with the configurations and number of objects, but if all scene instances share the same set of objects that follow the same physical laws, then transforming the problem of modeling scenes into a problem of modeling objects and the local physical processes that govern their interactions may provide a significant benefit in generalizing to solving novel physical tasks the learner has not encountered before. This is the central hypothesis of this paper. 

We test this hypothesis by defining models for perceiving and predicting raw observations that are themselves compositions of simpler functions that operate locally on \textit{entities} rather than globally on scenes.
Importantly, the symmetry that all objects follow the same physical laws enables us to define these learnable entity-centric functions to take as input argument a variable that represents a generic entity, the specific instantiations of which are all processed by the same function.
We use the term \textit{entity abstraction} to refer to the abstraction barrier that isolates the abstract \textit{variable}, which the entity-centric function is defined with respect to, from its concrete \textit{instantiation}, which contains information about the appearance and dynamics of an object that modulates the function's behavior.

\begin{figure}[!t]
\centering
\includegraphics[width=0.95\textwidth]{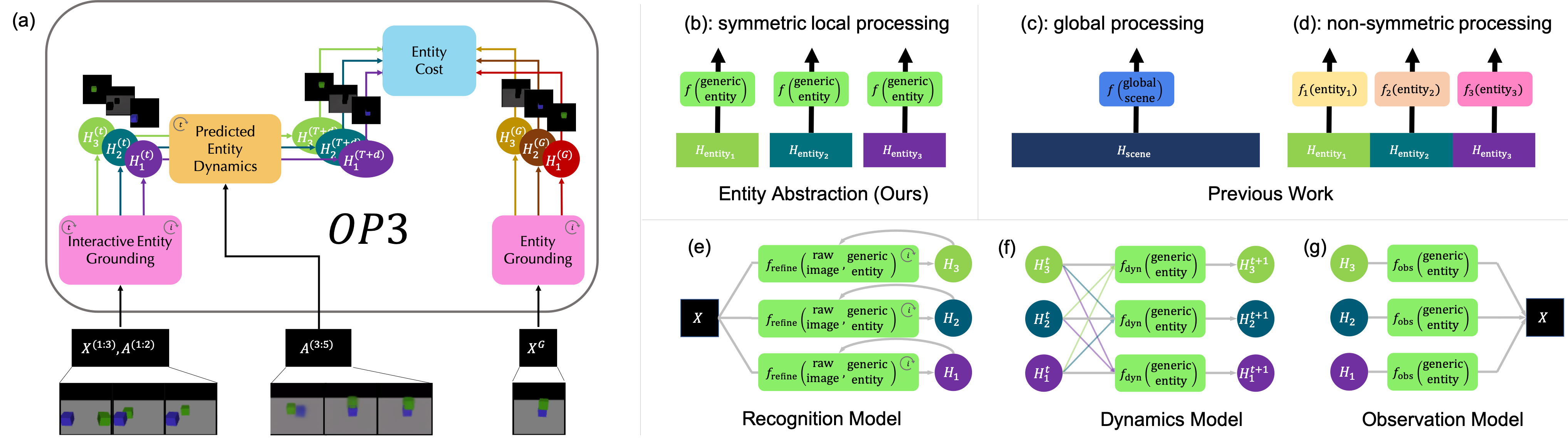}
\caption{\small{
\textbf{OP3}. \textbf{(a)} OP3 can infer a set of entity variables $H^{(T)}_{1:K}$ from a series of interactions (interactive entity grounding) or a single image (entity grounding). OP3 rollouts predict the future entity states $H^{(T+d)}_{1:K}$ given a sequence of actions $a^{(T:T+d)}$. We evaluate these rollouts during planning by scoring these predictions against inferred goal entity-states $H^{(G)}_k$.
\textbf{(b)} OP3 enforces the \textbf{entity abstraction}, factorizing the latent state into \textit{local} entity states, each of which are symmetrically processed with the same function that takes in a \textit{generic entity} as an argument.
In contrast, prior work either \textbf{(c)} process a global latent state~\citep{hafner2018learning} or \textbf{(d)} assume a fixed set of entities processed in a permutation-sensitive manner~\citep{finn2016unsupervised,kulkarni2019unsupervised,xu2018modeling,watters2019cobra}.
\textbf{(e-g)} Enforcing the entity-abstraction on modeling the \textbf{(f)} dynamics and \textbf{(g)} observation distributions of a POMDP, and on the \textbf{(e)} \textbf{interactive inference} procedure for grounding the entity variables in raw visual observations. Actions are not shown to reduce clutter.
}}
\label{fig:combined_overview}
\vspace{-10pt}
\end{figure}

Defining the observation and dynamic models of a model-based reinforcement learner as neural network functions of abstract entity variables allows for symbolic computation in the space of entities, but the key challenge for realizing this is to ground the values of these variables in the world from raw visual observations.
Fortunately, the language of partially observable Markov decision processes (POMDP) enables us to represent these entity variables as latent random state variables in a state-factorized POMDP, thereby transforming the variable binding problem into an inference problem with which we can build upon state-of-the-art techniques in amortized iterative variational inference~\citep{marino2018iterative,marino2018general,greff2019iodine} to use temporal continuity and interactive feedback to infer the posterior distribution of the entity variables given a sequence of observations and actions.

We present a framework for \textit{object-centric perception, prediction, and planning} (OP3), a model-based reinforcement learner that predicts and plans over entity variables inferred via an \textit{interactive inference} algorithm from raw visual observations.
Empirically OP3 learns to discover and bind information about actual objects in the environment to these entity variables \textit{without any supervision on what these variables should correspond to.}
As all computation within the entity-centric function is \textit{local in scope} with respect to its input entity, the process of modeling the dynamics or appearance of each object is \textit{protected} from the computations involved in modeling other objects, which allows OP3 to generalize to modeling a variable number of objects in a variety of contexts with no re-training.

\textbf{Contributions:} 
Our conceptual contribution is the use of entity abstraction to integrate graphical models, symbolic computation, and neural networks in a model-based reinforcement learning (RL) agent.
This is enabled by our technical contribution: defining models as the composition of locally-scoped entity-centric functions and the interactive inference algorithm for grounding the abstract entity variables in raw visual observations without any supervision on object identity.
Empirically, we find that OP3 achieves two to three times greater accuracy than state of the art video prediction models in solving novel single and multi-step block stacking tasks.
\section{Related Work} \label{sec:related_work}
\textbf{Representation learning for visual model-based reinforcement learning:}
Prior works have proposed learning video prediction models~\citep{wichers2018hierarchical,denton2017unsupervised,lee2018stochastic,finn2016unsupervised} to improve exploration~\citep{oh2015action} and planning~\citep{finn2017deep} in RL.
However, such works and others that represent the scene with a single representation vector~\citep{hafner2018learning,zhang2018solar,mnih-dqn-2015,oh2016control} may be susceptible to the binding problem~\citep{greff2015binding,rosenblatt1961principles} and must rely on data to learn that the same object in two different contexts can be modeled similarly.
But processing a disentangled latent state with a single function~\citep{whitney2016understanding,chen2016infogan,kulkarni2015deep,kulkarni2019unsupervised,goel2018unsupervised} or processing each disentangled factor in a permutation-sensitive manner~\citep{lee2018stochastic,xu2019unsupervised,kulkarni2019unsupervised} (1) assumes a fixed number of entities that cannot be dynamically adjusted for generalizing to more objects than in training and (2) has no constraints to enforce that multiple instances of the same entity in the scene be modeled in the same way.
For generalization, often the particular arrangement of objects in a scene does not matter so much as what is constant across scenes -- properties of individual objects and inter-object relationships -- which the inductive biases of these prior works do not capture.
The entity abstraction in OP3 enforces symmetric processing of entity representations, thereby overcoming the limitations of these prior works.

\textbf{Unsupervised grounding of abstract entity variables in concrete objects:}
Prior works that model entities and their interactions often pre-specify the identity of the entities~\citep{chang2016compositional,battaglia2016interaction,hamrick2017metacontrol,janner2018representation,narasimhan2018grounding,bapst2019structured,ajay2019combining}, provide additional supervision~\citep{girshick2014rich,he2017mask,wang2018deep,yang2018visual}, or provide additional specification such as segmentations~\citep{janner2018reasoning}, crops~\citep{fragkiadaki2015learning}, or a simulator~\citep{wu2017learning,kansky2017schema}.
Those that do not assume such additional information often factorize the entire scene into pixel-level entities~\citep{santoro2017simple,zambaldi2018deep,du2019task}, which do not model objects as coherent wholes.
None of these works solve the problem of grounding the entities in raw observation, which is crucial for autonomous learning and interaction.
OP3 builds upon recently proposed ideas in grounding entity representations via inference on a symmetrically factorized generative model of static~\citep{greff2015binding,greff2017neural,greff2019iodine} and dynamic~\citep{van2018relational} scenes, whose advantage over other methods for grounding~\citep{zhu2019object,eslami2016attend,burgess2019monet,kosiorek2018sequential,watters2019cobra} is the ability to refine the grounding with new information.
In contrast to other methods for binding in neural networks~\citep{levy2008vector,kanerva2009hyperdimensional,smolensky1990tensor,vaswani2017attention}, formulating inference as a mechanism for variable binding allows us to model uncertainty in the values of the variables.

\begin{figure}[!t]
    \centering
    \includegraphics[width=\textwidth]{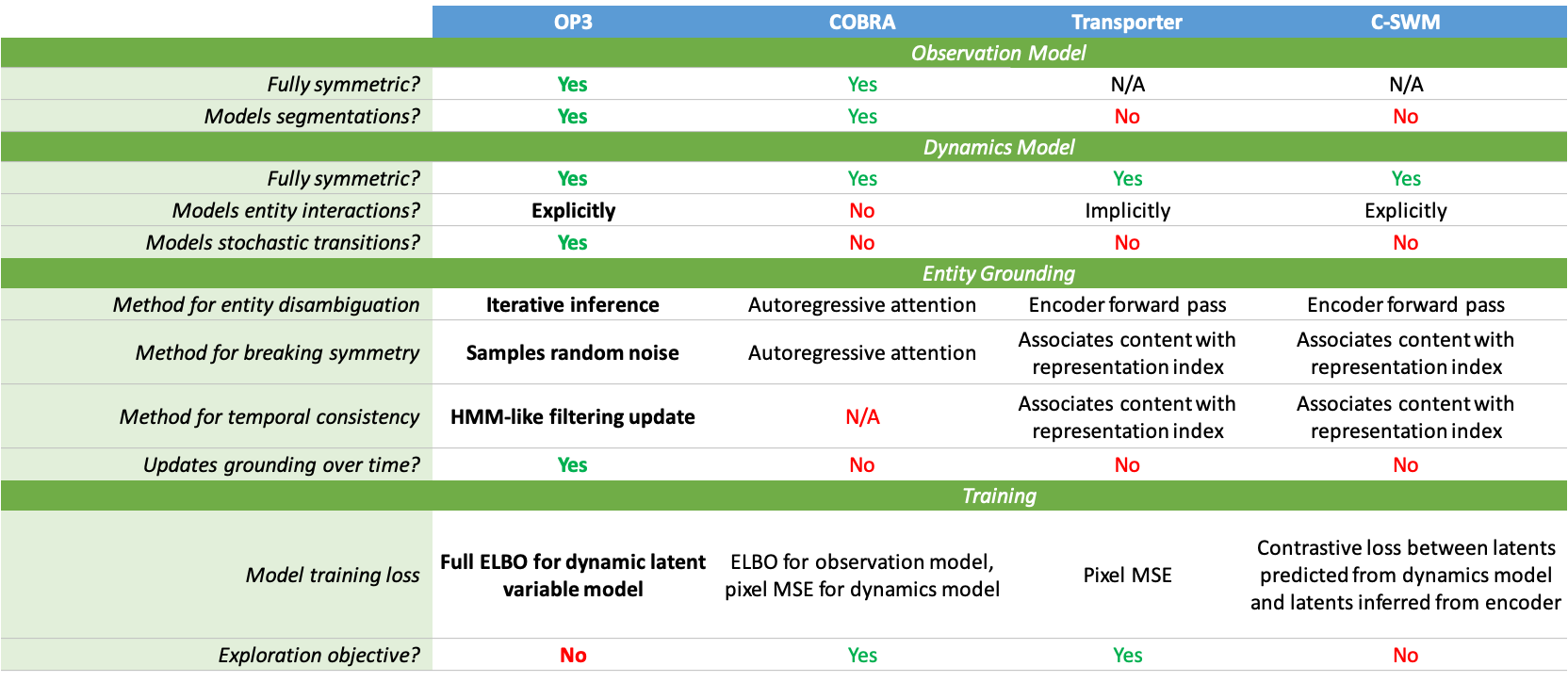}
    \caption{\small{
    \textbf{Comparison with other methods.}
    Unlike other methods, OP3 is a fully probabilistic factorized dynamic latent variable model, giving it several desirable properties.
    First, OP3 is naturally suited for combinatorial generalization~\citep{battaglia2018relational} because it enforces that local properties are invariant to changes in global structure.
    Because every learnable component of the OP3 operates symmetrically on each entity, including the mechanism that disambiguates entities itself (c.f. COBRA, which uses a learned autoregressive network to disambiguates entities, and Transporter and C-SWMs, which use a forward pass of a convolutional encoder for the global scene, rather than each entity), the weights of OP3 are invariant to changes in the number of instances of an entity, as well as the number of entities in the scene.
    Second, OP3's recurrent structure makes it straightforward to enforce spatiotemporal consistency, object permanence, and refine the grounding of its entity representations over time with new information. In contrast, COBRA, Transporter, and C-SWMs all model single-step dynamics and do not contain mechanisms for establishing a correspondence between the entity representations predicted from the previous timestep with the entity representations inferred at the current timestep.
    }}
    \label{fig:comparison table}
\end{figure}

\textbf{Comparison with similar work:}
The closest three works to OP3 are the Transporter~\citep{kulkarni2019unsupervised}, COBRA~\citep{watters2019cobra}, and C-SWMs~\citep{kipf2019contrastive}.
The Transporter enforces a sparsity bias to learn object keypoints, each represented as a feature vector at a pixel location, and the method's focus on keypoints has the advantage of enabling long-term object tracking, modeling articulated composite bodies such as joints, and scaling to dozens of objects.
C-SWMs learn entity representations using a contrastive loss, which has the advantage of overcoming the difficulty in attending to small but relevant features as well as the large model capacity requirements usually characteristic of the pixel reconstructive loss.
COBRA uses the autoregressive attention-based MONet~\citep{burgess2019monet} architecture to obtain entity representations, which has the advantage of being more computationally efficient and stable to train.
Unlike works such as \citep{greff2017neural,eslami2016attend,burgess2019monet,greff2019iodine} that infer entity representations from static scenes, these works represent complementary approaches to OP3 (Figure~\ref{fig:comparison table}) for representing dynamic scenes.

Symmetric processing of entities -- processing each entity representation with the same function, as OP3 does with its observation, dynamics, and refinement networks -- enforces the invariance that local properties are invariant to changes in global structure because it prevents the processing of one entity from being affected by other entities.
How symmetric the process is for obtaining these entity representations from visual observation affects how straightforward it is to directly transfer models of a single entity across different global contexts, such as in modeling multiple instances of the same entity in the scene in a similar way or in generalizing to modeling different numbers of objects than in training.
OP3 can exhibit this type of zero-shot transfer because the learnable components of its refinement process are fully symmetric across entities, which prevents OP3 from overfitting to the global structure of the scene.
In contrast, the KeyNet encoder of the Transporter and the CNN-encoder of C-SWMs associate the \textit{content} of the entity representation with the \textit{index} of that entity in a global representation vector (Figure~\ref{fig:combined_overview}d), and this permutation-sensitive mapping entangles the encoding of an entity with the global structure of the scene.
COBRA lies in between: it uses a learnable autoregressive attention network to infer segmentation masks, which entangles local object segmentations with global structure but may provide a useful bias for attending to salient objects, and symmetrically encodes entity representations given these masks\footnote{A discussion of the advantages and disadvantages of using an attention-based entity disambiguation method, which MONet and COBRA use, versus an iterative refinement method, which IODINE~\citep{greff2019iodine} and OP3 use, is discussed in~\citet{greff2019iodine}.}.

As a recurrent probabilistic dynamic latent variable model, OP3 can refine the grounding of its entity representations with new information from raw observations by simply applying a belief update similar to that used in filtering for hidden Markov models.
The Transporter, COBRA, and C-SWMs all do not have mechanisms for updating the belief of the entity representations with information from subsequent image frames.
Without recurrent structure, such methods rely on the assumption that a single forward pass of the encoder on a static image is sufficient to disambiguate objects, but this assumption is not generally true: objects can pop in and out of occlusion and what constitutes an object depends temporal cues, especially in real world settings.
Recurrent structure is built into the OP3 inference update (Appendix ~\ref{alg:appdx:timestep_t}), enabling OP3 to model object permanence under occlusion and refine its object representations with new information in modeling real world videos (Figure~\ref{fig:interactive_inference_real_world}).
\section{Problem Formulation} \label{sec:problem_formulation}
Let $x^*$ denote a physical scene and $h^*_{1:K}$ denote the objects in the scene.
Let $X$ and $A$ be random variables for the image observation of the scene $x^*$ and the agent's actions respectively.
In contrast to prior works~\citep{hafner2018learning} that use a single latent variable to represent the state of the scene, we use a set of latent random variables $H_{1:K}$ to represent the state of the objects $h^*_{1:K}$.
We use the term \textit{object} to refer to $h^*_k$, which is part of the physical world, and the term \textit{entity} to refer to $H_k$, which is part of our model of the physical world.
The generative distribution of observations $X^{(0:T)}$ and latent entities $H^{(0:T)}_{1:K}$ from taking $T$ actions $a^{(0:T-1)}$ is modeled as:
\footnotesize
\begin{equation} \label{eqn:generative}
    p\left(X^{(0:T)}, H^{(0:T)}_{1:K} \;\middle|\; a^{(0:T-1)}\right) = p\left(H^{(0)}_{1:K}\right) \prod_{t=1}^{T} p\left(H^{(t)}_{1:K} \;\middle|\; H^{(t-1)}_{1:K}, a^{(t-1)}\right) \prod_{t=0}^{T} p\left(X^{(t)} \;\middle|\; H^{(t)}_{1:K}\right)
\end{equation}
\normalsize
where $p(X^{(t)} \sgiven H^{(t)}_{1:K})$ and $p(H^{(t)}_{1:K} \sgiven H^{(t-1)}_{1:K}, A^{(t-1)})$ are the observation and dynamics distribution respectively shared across all timesteps $t$.
Our goal is to build a model that, from simply observing raw observations of random interactions, can generalize to solve novel compositional object manipulation problems that the learner was never trained to do, such as building various block towers during test time from only training to predict how blocks fall during training time.

When all tasks follow the same dynamics we can achieve such generalization with a planning algorithm if given a sequence of actions we could compute $p(X^{(T+1:T+d)} \sgiven X^{(0:T)}, A^{(0:T+d-1)})$, the posterior predictive distribution of observations $d$ steps into the future.
Approximating this predictive distribution can be cast as a variational inference problem (Appdx.~\ref{appdx:problem_formulation}) for learning the parameters of an approximate observation distribution $\mathdutchcal{G}(X^{(t)} \sgiven H^{(t)}_{1:K})$, dynamics distribution $\mathdutchcal{D}(H^{(t)}_{1:K} \sgiven H^{(t-1)}_{1:K}, A^{(t-1)})$, and a time-factorized recognition distribution $\mathdutchcal{Q}(H^{(t)}_{1:K} \sgiven H^{(t-1)}_{1:K}, X^{(t)}, A^{(t-1)})$ that maximize the evidence lower bound (ELBO), given by
$\mathcal{L} = \sum_{t=0}^{T} \mathcal{L}^{(t)}_{\text{r}} - \mathcal{L}^{(t)}_{\text{c}}$, where
\footnotesize
\begin{align*}
    \mathcal{L}^{t}_{\text{r}} &= \mathbb{E}_{h^t_{1:K} \sim q\left(H^{t}_{1:K} \sgiven h^{0:t-1}_{1:K}, x^{1:t}, a^{0:t-1} \right)}\left[\log \mathdutchcal{G}\left(x^t \sgiven h^t_{1:K}\right)\right]\\ 
    \mathcal{L}^{t}_{\text{c}} &= \mathbb{E}_{h^{t-1}_{1:K} \sim q\left(H^{t-1}_{1:K} \sgiven h^{1:t-2}_{1:K}, x^{1:t-1}, a^{0:t-2} \right)}\left[D_{KL}\left(\mathdutchcal{Q}\left(H^{t}_{1:K} \sgiven h^{t-1}_{1:K}, x^{t}, a^{t-1}\right) \kld \mathdutchcal{D}\left(H^{t}_{1:K} \sgiven h^{t-1}_{1:K}, a^{t-1} \right)\right)\right].
\end{align*}
\normalsize
The ELBO pushes $\mathdutchcal{Q}$ to produce states of the entities $H_{1:K}$ that contain information useful for not only reconstructing the observations via $\mathdutchcal{G}$ in $\mathcal{L}^{(t)}_{\text{r}}$ but also for predicting the entities' future states via $\mathdutchcal{D}$ in $\mathcal{L}^{(t)}_{\text{c}}$.
Sec.~\ref{sec:op3} will next offer our method for incorporating entity abstraction into modeling the generative distribution and optimizing the ELBO.

\section{Object-Centric Perception, Prediction, and Planning (OP3)} \label{sec:op3}
The \textit{entity abstraction} is derived from an assumption about symmetry: that the problem of modeling a dynamic scene of multiple entities can be reduced to the problem of (1) modeling a single entity and its interactions with an \textit{entity-centric} function and (2) applying this function to every entity in the scene.
Our choice to represent a scene as a set of entities exposes an avenue for directly encoding such a prior about symmetry that would otherwise not be straightforward with a global state representation.

As shown in Fig.~\ref{fig:combined_overview}, a function $F$ that respects the entity abstraction requires two ingredients.
The first ingredient (Sec.~\ref{sec:op3_model}) is that $F(H_{1:K})$ is expressed in part as the higher-order operation $\texttt{map}(f, H_{1:K})$ that broadcasts the same entity-centric function $f(H_k)$ to every entity variable $H_k$.
This yields the benefit of automatically transferring learned knowledge for modeling an individual entity to all entities in the scene rather than learn such symmetry from data.
As $f$ is a function that takes in a single generic entity variable $H_k$ as argument, the second ingredient (Sec.~\ref{sec:op3_inference}) should be a mechanism that binds information from the raw observation $X$ about a particular object $h^*_k$ to the variable $H_k$.

\subsection{Entity Abstraction in the Observation and Dynamics Models} \label{sec:op3_model}
The functions of interest in model-based RL are the observation and dynamics models $\mathdutchcal{G}$ and $\mathdutchcal{D}$ with which we seek to approximate the data-generating distribution in equation~\ref{eqn:generative}.

\begin{wrapfigure}{R}{0.45\textwidth}
\centering
\includegraphics[width=0.45\textwidth]{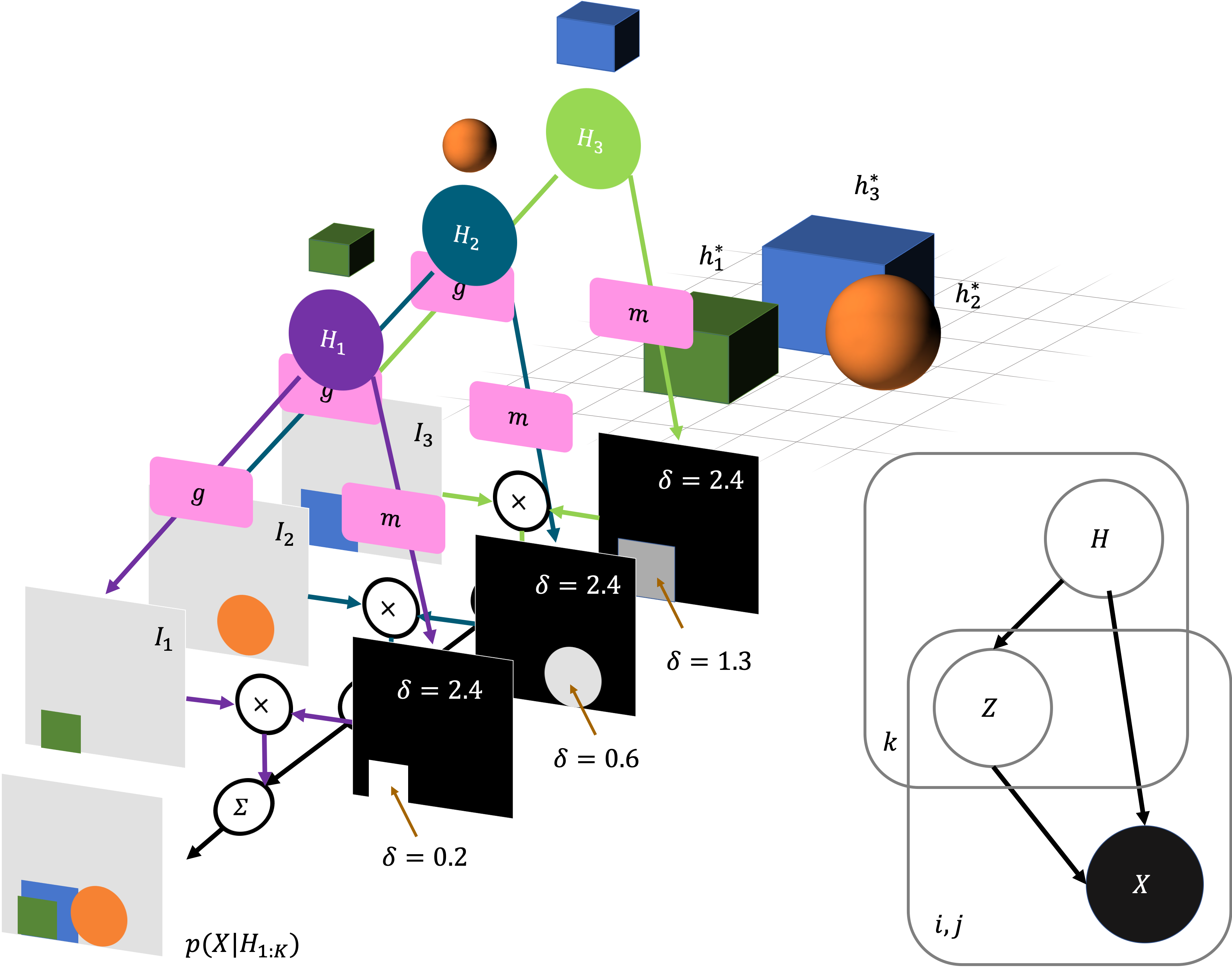}
\caption{\small{
\textbf{(a)} The observation model $\mathdutchcal{G}$ models an observation image as a composition of sub-images weighted by segmentation masks. The shades of gray in the masks indicate the depth $\delta$ from the camera of the object that the sub-image depicts.
\textbf{(b)} The graphical model of the generative model of observations, where $k$ indexes the entity, and $i,j$ indexes the pixel.
$Z$ is the indicator variable that signifies whether an object's depth at a pixel is the closest to the camera.
}}
\label{fig:observation_model}
\end{wrapfigure}

\textbf{Observation Model:} 
The observation model $\mathdutchcal{G}(X \sgiven H_{1:K})$ approximates the distribution $p(X \sgiven H_{1:K})$, which models how the observation $X$ is caused by the combination of entities $H_{1:K}$.
We enforce the entity abstraction in $\mathdutchcal{G}$ (in Fig.~\ref{fig:combined_overview}g) by applying the same entity-centric function $\mathdutchcal{g}(X \sgiven H_k)$ to each entity $H_k$, which we can implement using a mixture model at each pixel $(i,j)$:
\footnotesize
\begin{equation} \label{eqn:log_likelihood}
    \mathdutchcal{G}\left(X_{(ij)} \given H_{1:K}\right) = \sum_{k=1}^K m_{(ij)}\left(H_k\right) \cdot \mathdutchcal{g}\left(X_{(ij)} \sgiven H_k\right),
\end{equation}
\normalsize
where $\mathdutchcal{g}$ computes the mixture components that model how each individual entity $H_k$ is independently generated, combined via mixture weights $m$ that model the entities' relative depth from the camera, the derivation of which is in Appdx.~\ref{appdx:obs_model}.

\textbf{Dynamics Model:} The dynamics model $\mathdutchcal{D}(H'_{1:K} \sgiven H_{1:K}, A)$ approximates the distribution $p(H'_{1:K} \sgiven H_{1:K}, A)$, which models how an action $A$ intervenes on the entities $H_{1:K}$ to produce their future values $H'_{1:K}$.
We enforce the entity abstraction in $\mathdutchcal{D}$ (in Fig.~\ref{fig:combined_overview}f) by applying the same entity-centric function $\mathdutchcal{d}(H'_k \sgiven H_k, H_{[\neq k]}, A)$ to each entity $H_k$, which reduces the problem of modeling how an action affects a scene with a combinatorially large space of object configurations to the problem of simply modeling how an action affects a single \textit{generic} entity $H_k$ and its interactions with the list of other entities $H_{[\neq k]}$.
Modeling the action as an finer-grained intervention on a \textit{single} entity rather than the entire scene is a benefit of using local representations of entities rather than global representations of scenes.

However, at this point we still have to model the combinatorially large space of \textit{interactions} that a single entity could participate in.
Therefore, we can further enforce a \textit{pairwise} entity abstraction on $\mathdutchcal{d}$ by applying the same \textit{pairwise} function $\mathdutchcal{d}_{oo}(H_k, H_i)$ to each entity pair $(H_k, H_i)$, for $i \in [\neq k]$.
Omitting the action to reduce clutter (the full form is written in Appdx.~\ref{appdx:dyn_model}), the structure of the $\mathdutchcal{D}$ therefore follows this form:
\footnotesize
\begin{align}
    \mathdutchcal{D}\left(H'_{1:K} \given H_{1:K}\right) = \prod_{k=1}^K \mathdutchcal{d}\left(H'_k \given H_k, H^{\text{interact}}_{k}\right), \;\text{where}\; H_k^{\text{interact}} = \sum_{i \neq k}^K d_{oo}\left(H_i, H_k\right).
\end{align}
\normalsize
The entity abstraction therefore provides the flexibility to scale to modeling a variable number of objects by solely learning a function $\mathdutchcal{d}$ that operates on a single generic entity and a function $\mathdutchcal{d}_{oo}$ that operates on a single generic entity pair, both of which can be re-used for across all entity instances.

\begin{figure}
\centering
\includegraphics[width=0.95\textwidth]{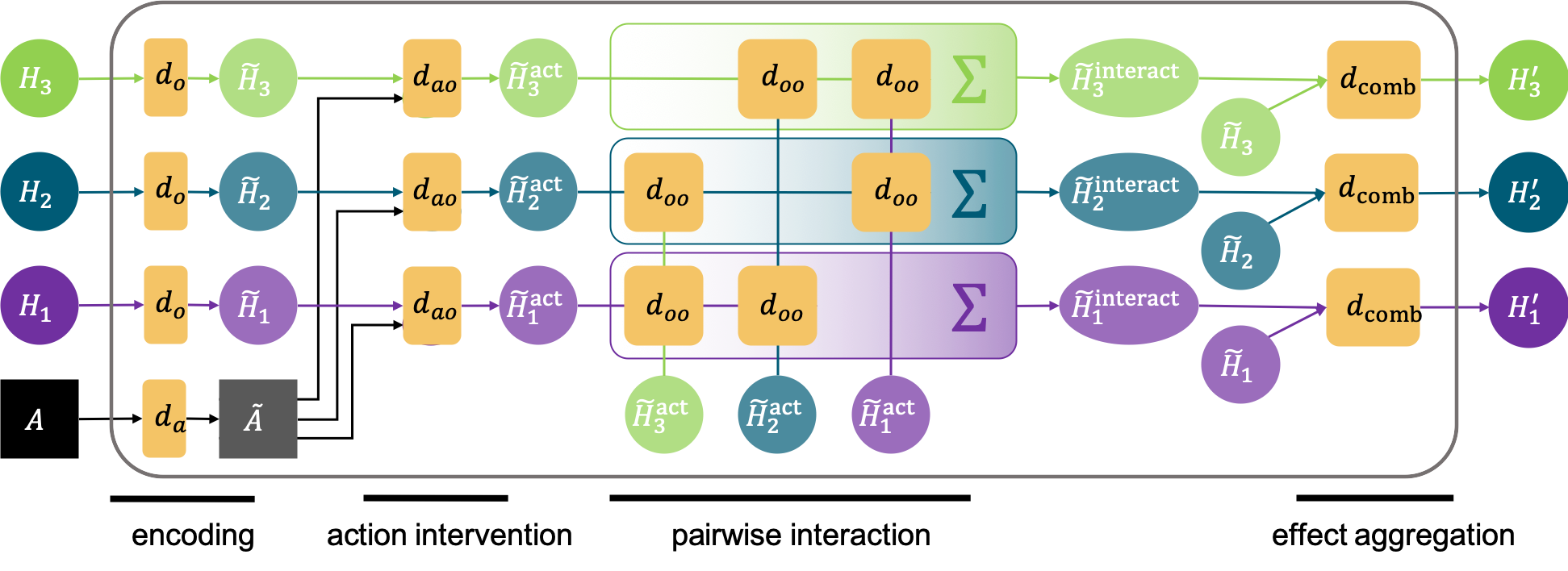}
\caption{\small{The \textbf{dynamics model} $\mathdutchcal{D}$ models the time evolution of every object by symmetrically applying the function $\mathdutchcal{d}$ to each object.
For a given object, $d$ models the individual dynamics of that object $\left(d_o\right)$, embeds the action vector $\left(d_a\right)$, computes the action's effect on that object $\left(d_{ao}\right)$, computes each of the other objects' effect on that object $\left(d_{oo}\right)$, and aggregates these effects together $\left(d_{\text{comb}}\right)$.
}}
\label{fig:dynamics}
\end{figure}
\subsection{Interactive Inference for Binding Object Properties to Latent Variables} \label{sec:op3_inference}
For the observation and dynamics models to operate from raw pixels hinges on the ability to bind the properties of specific physical objects $h^*_{1:K}$ to the entity variables $H_{1:K}$.
For latent variable models, we frame this variable binding problem as an inference problem: binding information about $h^*_{1:K}$ to $H_{1:K}$ can be cast as a problem of inferring the parameters of $p(H^{(0:T)} \sgiven x^{(0:T)}, a^{(0:T-1)})$, the posterior distribution of $H_{1:K}$ given a sequence of interactions.
Maximizing the ELBO in Sec.~\ref{sec:problem_formulation} offers a method for learning the parameters of the observation and dynamics models while simultaneously learning an approximation to the posterior $q(H^{(0:T)} \sgiven x^{(0:T)}, a^{(0:T-1)}) = \prod_{t=0}^T \mathdutchcal{Q}(H^{(t)}_{1:K} \sgiven H^{(t-1)}_{1:K}, x^{(t)}, a^{(t)})$, which we have chosen to factorize into a per-timestep recognition distribution $\mathdutchcal{Q}$ shared across timesteps.
We also choose to enforce the entity abstraction on the process that computes the recognition distribution $\mathdutchcal{Q}$ (in Fig.~\ref{fig:combined_overview}e) by decomposing it into a recognition distribution $\mathdutchcal{q}$ applied to each entity:
\footnotesize
\begin{align}
    \mathdutchcal{Q}\left(H^{(t)}_{1:K} \sgiven h^{(t-1)}_{1:K}, x^{(t)}, a^{(t)}\right) = \prod_{k=1}^K \mathdutchcal{q}\left(H^{(t)}_{k} \sgiven h^{(t-1)}_{k}, x^{(t)}, a^{(t)}\right).
\end{align} 
\normalsize
Whereas a neural network encoder is often used to approximate the posterior~\citep{hafner2018learning,xu2018modeling,kulkarni2019unsupervised}, a single forward pass that computes $\mathdutchcal{q}$ in parallel for each entity is insufficient to break the symmetry for dividing responsibility of modeling different objects among the entity variables~\citep{zhang2019set} because the entities do not have the opportunity to communicate about which part of the scene they are representing.

We therefore adopt an \textit{iterative inference} approach~\citep{marino2018iterative} to compute the recognition distribution $\mathdutchcal{Q}$, which has been shown to break symmetry among modeling objects in static scenes~\citep{greff2019iodine}.
Iterative inference computes the recognition distribution via a \textit{procedure}, rather than a single forward pass of an encoder, that iteratively refines an initial guess for the posterior parameters $\lambda_{1:K}$ by using gradients from how well the generative model is able to predict the observation based on the current posterior estimate.
The initial guess provides the noise to break the symmetry.

\begin{figure}
\centering
\includegraphics[width=0.9\textwidth]{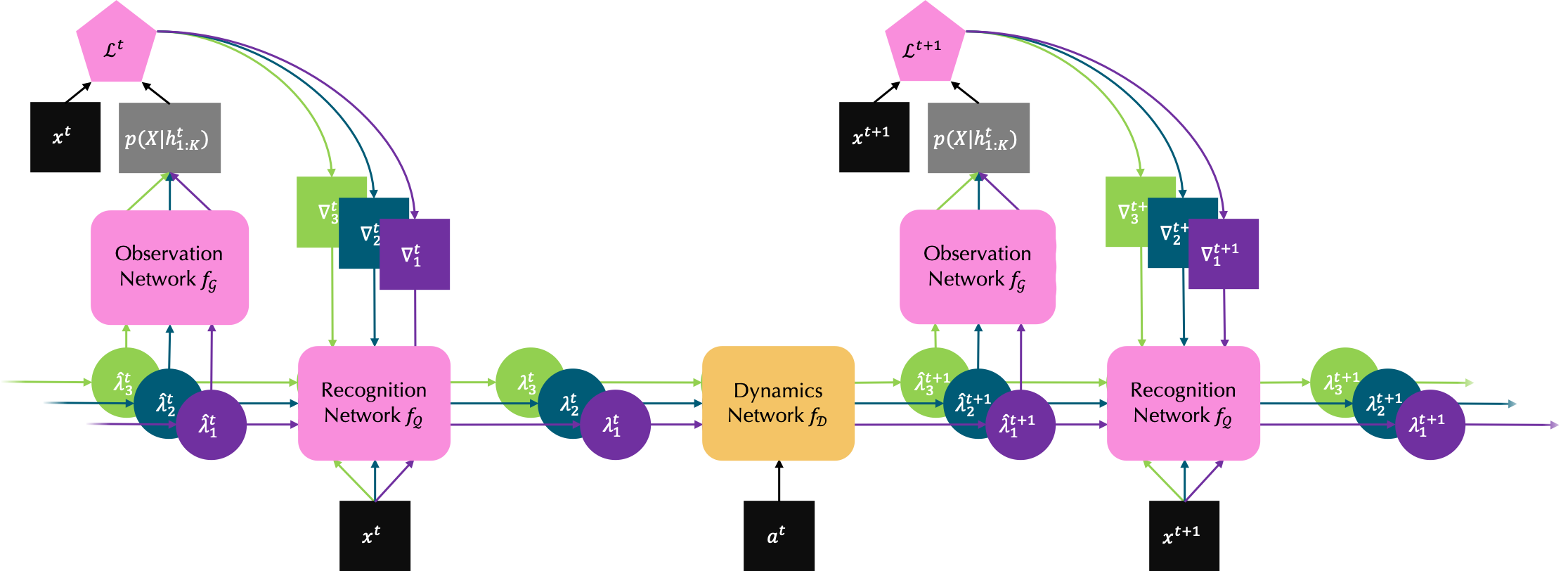}
\caption{\small{
\textbf{Amortized interactive inference} alternates between refinement (pink) and dynamics (orange) steps, iteratively updating the belief of $\lambda_{1:K}$ over time.
$\hat{\lambda}$ corresponds to the output of the dynamics network, which serves as the initial estimate of $\lambda$ that is subsequently refined by $f_\mathdutchcal{G}$ and $f_\mathdutchcal{Q}$. $\triangledown$ denotes the feedback used in the refinement process, which includes gradient information and auxiliary inputs (Appdx.~\ref{appdx:interactive_inference}).
}}
\label{fig:inference}
\end{figure}

For scenes where position and color are enough for disambiguating objects, a static image may be sufficient for inferring $\mathdutchcal{q}$.
However, in interactive environments disambiguating objects is more underconstrained because what constitutes an object depends on the goals of the agent.
We therefore incorporate actions into the amortized varitional filtering framework~\citep{marino2018general} to develop an \textit{interactive inference} algorithm (Appdx.~\ref{appdx:interactive_inference} and Fig.~\ref{fig:inference}) that uses temporal continuity and interactive feedback to disambiguate objects.
Another benefit of enforcing entity abstraction is that preserving temporal consistency on entities comes for free: information about each object remains bound to its respective $H_k$ through time, mixing with information about other entities only through explicitly defined avenues, such as in the dynamics model.

\subsection{Training at Different Timescales} \label{sec:training}
The variational parameters $\lambda_{1:K}$ are the interface through which the neural networks $f_{\mathdutchcal{g}}$, $f_{\mathdutchcal{d}}$, $f_{\mathdutchcal{q}}$ that respectively output the distribution parameters of $\mathdutchcal{G}$, $\mathdutchcal{D}$, and $\mathdutchcal{Q}$ communicate.
For a \textit{particular} dynamic scene, the execution of interactive inference optimizes the variational parameters $\lambda_{1:K}$.
\textit{Across} scene instances, we train the weights of $f_{\mathdutchcal{g}}$, $f_{\mathdutchcal{d}}$, $f_{\mathdutchcal{q}}$ by backpropagating the ELBO through the entire inference procedure, spanning multiple timesteps.
OP3 thus learns at three different timescales: the variational parameters learn (1) across $M$ steps of inference within a single timestep and (2) across $T$ timesteps within a scene instance, and the network weights learn (3) across different scene instances.

Beyond next-step prediction, we can directly train to compute the posterior predictive distribution $p(X^{(T+1:T+d)} \sgiven x^{(0:T)}, a^{(0:T+d)})$ by sampling from the approximate posterior of $H_{1:K}^{(T)}$ with $\mathdutchcal{Q}$, rolling out the dynamics model $\mathdutchcal{D}$ in latent space from these samples with a sequence of $d$ actions, and predicting the observation $X^{(T+d)}$ with the observation model $\mathdutchcal{G}$.
This approach to action-conditioned video prediction predicts future observations directly from observations and actions, but with a bottleneck of $K$ time-persistent entity-variables with which the dynamics model $\mathdutchcal{D}$ performs symbolic relational computation.

\subsection{Object-Centric Planning} \label{sec:planning}
OP3 rollouts, computed as the posterior predictive distribution, can be integrated into the standard visual model-predictive control~\citep{finn2017deep} framework.
Since interactive inference grounds the entities $H_{1:K}$ in the actual objects $h_{1:K}^*$ depicted in the raw observation, this grounding essentially gives OP3 access to a \textit{pointer} to each object, enabling the rollouts to be in the space of entities and their relations.
These pointers enable OP3 to not merely predict in the space of entities, but give OP3 access to an \textit{object-centric action space}: 
for example, instead of being restricted to the standard \texttt{(pick\_xy, place\_xy)} action space common to many manipulation tasks, which often requires biased picking with a scripted policy~\citep{levine2018learning,kalashnikov2018qt}, these pointers enable us to compute a mapping (Appdx.~\ref{appdx:multi_step_block_stacking}) between \texttt{entity\_id} and \texttt{pick\_xy}, allowing OP3 to automatically use a \texttt{(entity\_id, place\_xy)} action space without needing a scripted policy.

\subsection{Generalization to Various Tasks} \label{sec:cost_function}
We consider tasks defined in the same environment with the same physical laws that govern appearance and dynamics.
Tasks are differentiated by goals, in particular goal configurations of objects.
Building good cost functions for real world tasks is generally difficult~\citep{fu2018variational} because the underlying state of the environment is always unobserved and can only be modeled through modeling observations.
However, by representing the environment state as the state of its entities, we may obtain finer-grained goal-specification without the need for manual annotations~\citep{ebert2018visual}.
Having rolled out OP3 to a particular timestep, we construct a cost function to compare the predicted entity states $H_{1:K}^{(P)}$ with the entity states $H_{1:K}^{(G)}$ inferred from a goal image by considering pairwise distances between the entities, another example of enforcing the pairwise entity abstraction.
Letting $S'$ and $S$ denote the set of goal and predicted entities respectively, we define the form of the cost function via a composition of the task specific distance function $\mathdutchcal{c}$ operating on entity-pairs:
\footnotesize
\begin{equation}
    \mathdutchcal{C}\left(H_{1:K}^{(G)}, H_{1:K}^{(P)}\right) = \sum_{a \in S'} \min_{b \in S} \; \mathdutchcal{c}\left(H_{a}^{(G)}, H_{b}^{(P)}\right),
\end{equation}
\normalsize
in which we pair each goal entity with the closest predicted entity and sum over the costs of these pairs.
Assuming a single action suffices to move an object to its desired goal position, we can greedily plan each timestep by defining the cost to be $\min_{a \in S', b \in S} \; \mathdutchcal{c}(H_{a}^{(G)}, H_{b}^{(P)})$, the pair with minimum distance, and removing the corresponding goal entity from further consideration for future planning.
\section{Experiments} \label{sec:experiments}
Our experiments aim to 
study to what degree entity 
abstraction improves generalization, planning, and modeling.
Sec.~\ref{sec:single_step} shows that from only training to predict how objects fall, OP3 generalizes to solve various novel block stacking tasks with two to three times better accuracy than a state-of-the-art video prediction model. Sec.~\ref{sec:multi_step} shows that OP3 can plan for multiple steps in a difficult multi-object environment. Sec.~\ref{sec:real_world} shows that OP3 learns to ground its abstract entities in objects from real world videos.

\subsection{Combinatorial Generalization without Object Supervision} \label{sec:single_step}

We first investigate how well OP3 can learn object-based representations without additional object supervision, as well as how well OP3's factorized representation can enable combinatorial generalization for scenes with many objects.

\textbf{Domain:} In the MuJoCo~\citep{todorov2012mujoco} block stacking task introduced by~\citet{janner2018reasoning} for the O2P2 model, a block is raised in the air and the model must  predict the steady-state effects of dropping the block on a surface with multiple objects, which implicitly requires modeling the effects of gravity and collisions. The agent is never trained to stack blocks, but is tested on a suite of tasks where it must construct block tower specified by a goal image. \citet{janner2018reasoning} showed that an object-centric model with access to \textit{ground truth} object segmentations can solve these tasks with about 76\% accuracy. We now consider whether OP3 can do better, but \textit{without any supervision on object identity}.

\begin{minipage}{\textwidth}
\begin{minipage}[m]{.39\textwidth}
\footnotesize
    \centering
    \begin{tabular}{ccc}
    \toprule
    SAVP   &  O2P2 &  OP3 (ours)   \\
    \midrule
    24\% & 76\% & \textbf{82\%}\\
    \bottomrule
    \end{tabular}
    \centering
    \captionof{table}{\small{Accuracy (\%) of block tower builds by the SAVP baseline, the O2P2 oracle, and our approach. O2P2 uses image segmentations whereas OP3 uses only raw images as input.}}
    \label{tab:mse_results}
\end{minipage}
 \hfill
\begin{minipage}[m]{.59\textwidth}
\footnotesize
\centering
            \begin{tabular}{c|ccc}
    \toprule
            \# Blocks & SAVP    &  OP3 (xy) & OP3 (entity) \\
         \midrule
              1 & 54\% & 73\% & \textbf{91\%}\\
              2 & 28\%  & 55\% & \textbf{80\%}\\
               3 & 28\%  & 41\% & \textbf{55\%}\\
    \bottomrule
    \end{tabular}
    \centering
    \captionof{table}{\small{Accuracy (\%) of multi-step planning for building block towers. (xy) means \texttt{(pick\_xy, place\_xy)} action space while (entity) means \texttt{(entity\_id, place\_xy)} action space.}}
    \label{tab:multistep_results}
\end{minipage}
\normalsize
\end{minipage}

\textbf{Setup:} We train OP3 on the same dataset and evaluate on the same goal images as~\citet{janner2018reasoning}.
While the training set contains up to five
objects, the test set contains up to nine objects, which are placed in specific structures (bridge, pyramid, etc.) not seen during training.
The actions are optimized using the cross-entropy method (CEM)~\citep{Rubinstein2004TheCM}, with each sampled action evaluated by the greedy cost function described in Sec.~\ref{sec:cost_function}.
Accuracy is evaluated using the metric defined by \citet{janner2018reasoning}, which checks that all blocks are within some threshold error of the goal.

\textbf{Results:} 
The two baselines, SAVP~\citep{lee2018stochastic} and O2P2, represent the state-of-the-art in video prediction and symmetric object-centric planning methods, respectively.
SAVP models objects with a fixed number of convolutional filters and does not process entities symmetrically.
O2P2 does process entities symmetrically, but requires access to ground truth object segmentations.
As shown in Table~\ref{tab:mse_results}, OP3 achieves better accuracy than O2P2, even without any ground truth supervision on object identity, possibly because grounding the entities in the raw image may provide a richer contextual representation than encoding each entity separately without such global context as O2P2 does.
OP3 achieves three times the accuracy of SAVP, which suggests that symmetric modeling of entities is enables the flexibility to transfer knowledge of dynamics of a single object to novel scenes with different configurations heights, color combinations, and numbers of objects than those from the training distribution.
Fig.~\ref{fig:mpc_results} and Fig.~\ref{fig:predicted_components} in the Appendix show that, by grounding its entities in objects of the scene through inference, OP3's predictions isolates only one object at a time without affecting the predictions of other objects.
 
\begin{figure}[!h]
\centering
\includegraphics[width=\textwidth]{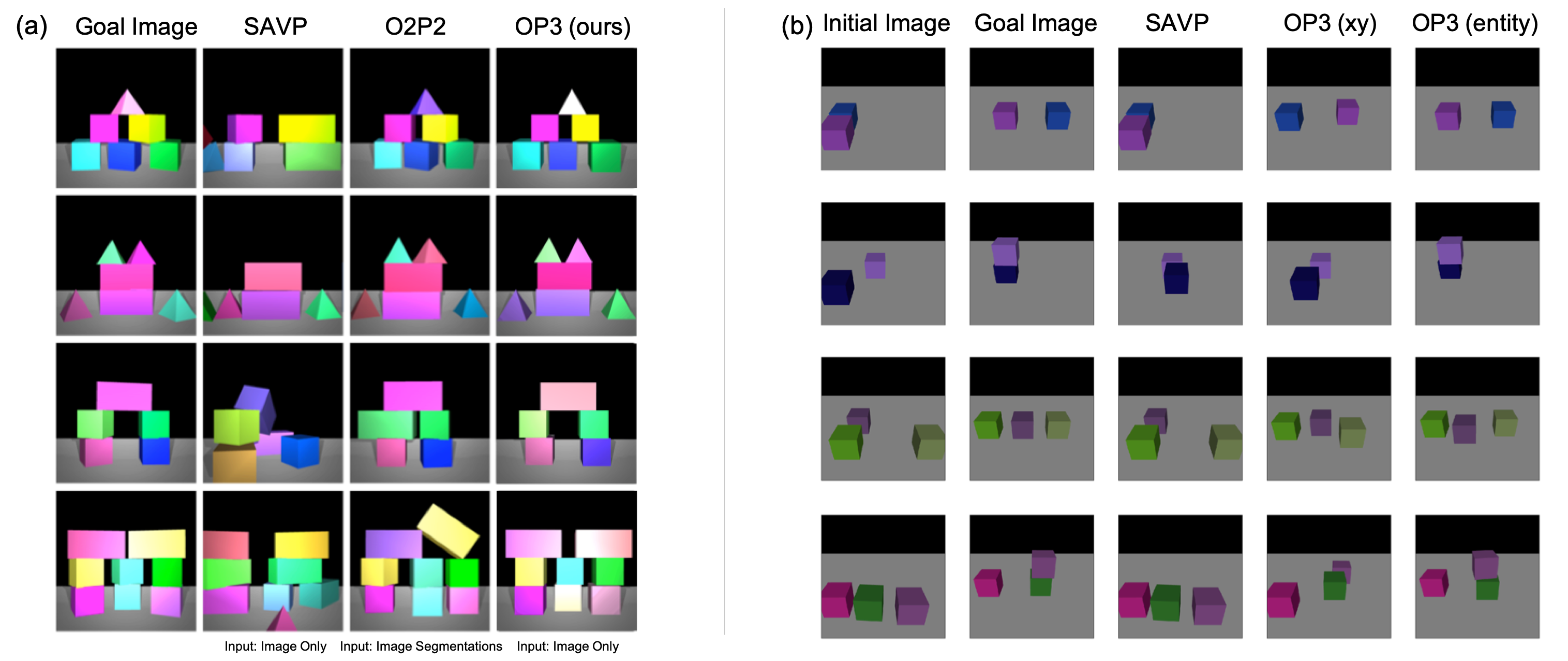}
\caption{\small{
\textbf{(a)} In the block stacking task from~\citep{janner2018reasoning} with single-step greedy planning, OP3's generalizes better than both O2P2, an oracle model with access to image segmentations, and SAVP, which does not enforce entity abstraction.
\textbf{(b)} OP3 exhibits better multi-step planning with objects already present in the scene. By planning with MPC using random pick locations (SAVP and OP3 (xy)), the sparsity of objects in the scene make it rare for random pick locations to actually pick the objects. However, because OP3 has access to pointers to the latent entities, we can use these to automatically bias the pick locations to be at the object location, without any supervision (OP3 (entity)).
}}
\label{fig:stage_1_stage_3}
\end{figure}

\subsection{Multi-Step Planning} \label{sec:multi_step}
The goal of our second experiment is to understand how well OP3 can perform multi-step planning by manipulating objects already present in the scene. We modify the block stacking task by changing the action space to represent a picking and dropping location. This requires reasoning over extended action sequences since moving objects out of place may be necessary.

Goals are specified with a goal image, and the initial scene contains all of the blocks needed to build the desired structure. This task is more difficult because the agent may have to move blocks out of the way before placing other ones which would require multi-step planning. Furthermore, an action only successfully picks up a block if it intersects with the block's outline, which makes searching through the combinatorial space of plans a challenge. As stated in Sec.~\ref{sec:planning}, having a pointer to each object enables OP3 to plan in the space of entities.
We compare two different action spaces \texttt{(pick\_xy, place\_xy)} and \texttt{(entity\_id, place\_xy)} to understand how automatically filtering for pick locations at actual locations of objects enables better efficiency and  performance in planning.
Details for determining the \texttt{pick\_xy} from \texttt{entity\_id} are in appendix \ref{appdx:multi_step_block_stacking}.

\begin{figure}[!h]
\centering
\includegraphics[width=\textwidth]{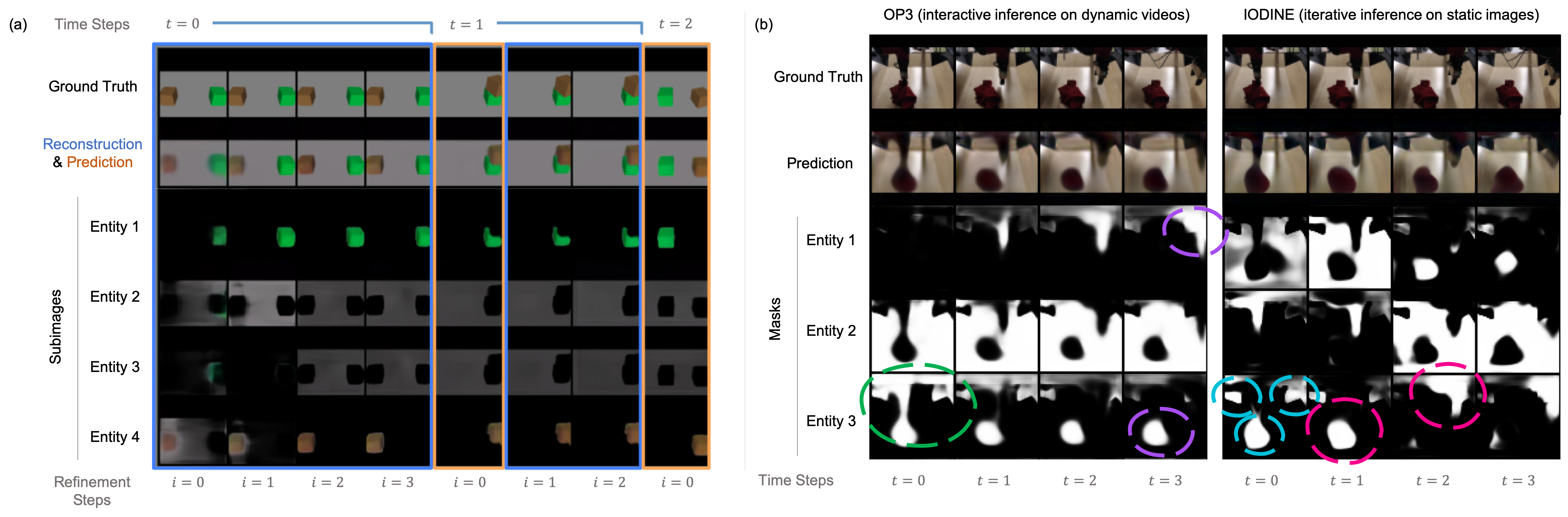}
\caption{\small{
Visualization of interactive inference for block-manipulation and real-world videos~\citep{ebert2018robustness}. Here, OP3 interacts with the objects by executing pre-specified actions in order to disambiguate objects already present in the scene by taking advantage of temporal continuity and receiving feedback from how well its prediction of how an action affects an object compares with the ground truth result.
\textbf{(a)} OP3 does four refinement steps on the first image, and then 2 refinement steps after each prediction.
\textbf{(b)} We compare OP3, applied on dynamic videos, with IODINE, applied independently to each frame of the video, to illustrate that using a dynamics model to propagate information across time enables better object disambiguation. We observe that initially, both OP3 (green circle) and IODINE (cyan circles) both disambiguate objects via color segmentation because color is the only signal in a static image to group pixels. However, we observe that as time progresses, OP3 separates the arm, object, and background into separate latents (purple) by using its currently estimates latents predict the next observation and comparing this prediction with the actually observed next observation. In contrast, applying IODINE on a per-frame basis does not yield benefits of temporal consistency and interactive feedback (red).
}}
\label{fig:interactive_inference_real_world}
\end{figure}

\textbf{Results:} We compare with SAVP, which uses the \texttt{(pick\_xy, place\_xy)} action space.
With this standard action space (Table \ref{tab:multistep_results}) OP3 achieves between 1.5-2 times the accuracy of SAVP. This performance gap increases to 2-3 times the accuracy when OP3 uses the \texttt{(entity\_id, place\_xy)} action space.
The low performance of SAVP with only two blocks highlights the difficulty of such combinatorial tasks for model-based RL methods, and highlights the both the generalization and localization benefits of a model with entity abstraction.
Fig.~\ref{fig:stage_1_stage_3}b shows that OP3 is able to plan more efficiently, suggesting that OP3 may be a more effective model than SAVP in modeling combinatorial scenes.
Fig.~\ref{fig:interactive_inference_real_world}a shows the execution of interactive inference during training, where OP3 alternates between four refinement steps and one prediction step.
Notice that OP3 infers entity representations that decompose the scene into coherent objects and that entities that do not model objects model the background.
We also observe in the last column ($t=2$) that OP3 predicts the appearance of the green block even though the green block was partially occluded in the previous timestep, which shows its ability to retain information across time.

\subsection{Real World Evaluation} \label{sec:real_world}
The previous tasks used simulated environments with monochromatic objects. Now we study how well OP3 scales to real world data with cluttered scenes, object ambiguity, and occlusions. We evaluate OP3 on the dataset from~\citet{ebert2018robustness} which contains videos of a robotic arm moving cloths and other deformable and multipart objects with varying textures.

We evaluate qualitative performance by visualizing the object segmentations and compare against vanilla IODINE, which does not incorporate an interaction-based dynamics model into the inference process. Fig.~\ref{fig:interactive_inference_real_world}b highlights the strength of OP3 in preserving temporal continuity and disambiguating objects in real world scenes. While IODINE can disambiguate monochromatic objects in static images, we observe that it struggles to do more than just color segmentation on more complicated images where movement is required to disambiguate objects.
In contrast, OP3 is able to use temporal information to obtain more accurate segmentations, as seen in Fig.~\ref{fig:interactive_inference_real_world}b where it initially performs color segmentation by grouping the towel, arm, and dark container edges together, and then by observing the effects of moving the arm, separates these entities into different groups.

\section{Discussion}
We have shown that enforcing the entity abstraction in a model-based reinforcement learner improves generalization, planning, and modeling across various compositional multi-object tasks.
In particular, enforcing the entity abstraction provides the learner with a pointer to each entity variable, enabling us to define functions that are local in scope with respect to a particular entity, allowing knowledge about an entity in one context to directly transfer to modeling the same entity in different contexts.
In the physical world, entities are often manifested as objects, and generalization in physical tasks such as robotic manipulation often may require symbolic reasoning about objects and their interactions.
However, the general difficulty with using purely symbolic, abstract representations is that it is unclear how to continuously update these representations with more raw data.
OP3 frames such symbolic entities as random variables in a dynamic latent variable model and infers and refines the posterior of these entities over time with neural networks.
This suggests a potential bridge to connect abstract symbolic variables with the noisy, continuous, high-dimensional physical world, opening a path to scaling robotic learning to more combinatorially complex tasks.



\acknowledgments{The authors would like to thank the anonymous reviewers for their helpful feedback and comments.
The authors would also like to thank Sjoerd van Steenkiste, Nalini Singh and Marvin Zhang for helpful discussions on the graphical model, Klaus Greff for help in implementing IODINE, Alex Lee for help in running SAVP, Tom Griffiths, Karl Persch, and Oleg Rybkin for feedback on earlier drafts, Joe Marino for discussions on iterative inference, and Sam Toyer, Anirudh Goyal, Jessica Hamrick, Peter Battaglia, Loic Matthey, Yash Sharma, and Gary Marcus for insightful discussions.
This research was supported in part by the National Science Foundation under IIS-1651843, IIS-1700697, and IIS-1700696, the Office of Naval Research, ARL DCIST CRA W911NF-17-2-0181, DARPA, Berkeley DeepDrive, Google, Amazon, and NVIDIA.}
\clearpage

\scriptsize
\bibliography{references}  
\normalsize
\newpage

\appendix
\setlength{\abovedisplayskip}{2pt}
\setlength{\belowdisplayskip}{2pt}

\section{Observation Model} \label{appdx:obs_model}
The observation model $\mathdutchcal{G}$ models how the objects $H_{1:K}$ cause the image observation $X \in \mathbbm{R}^{N \times M}$.
Here we provide a mechanistic justification for our choice of observation model by formulating the observation model as a probabilistic approximation to a deterministic rendering engine.

\textbf{Deterministic rendering engine:} Each object $H_k$ is rendered independently as the sub-image $I_k$ and the resulting $K$ sub-images are combined to form the final image observation $X$. To combine the sub-images, each pixel $I_{k(ij)}$ in each sub-image is assigned a depth $\delta_{k(ij)}$ that specifies the distance of object $k$ from the camera at coordinate $(ij)$. of the image plane. Thus the pixel $X_{(ij)}$ takes on the value of its corresponding pixel $I_{k(ij)}$ in the sub-image $I_k$ if object $k$ is closest to the camera than the other objects, such that
\footnotesize
\begin{equation}
    X_{(ij)} = \sum_{k=1}^K Z_{k(ij)} \cdot I_{k(ij)},
\end{equation}
\normalsize
where $Z_{k(ij)}$ is the indicator random variable $\mathbbm{1}[k = \argmin_{k \in K} \delta_{k(ij)}]$, allowing us to intuitively interpret $Z_{k}$ as segmentation masks and $I_{k}$ as color maps.

\textbf{Modeling uncertainty with the observation model:}
In reality we do not directly observe the depth values, so we must construct a probabilistic model to model our uncertainty:
\footnotesize
\begin{equation} \label{eqn:qppdx:obs_model}
    \mathdutchcal{G}\left(X| H_{1:K}\right) = \prod_{i,j=1}^{N,M} \sum_{k=1}^K m_{(ij)}(H_k) \cdot \mathdutchcal{g}\left(X_{(ij)} \sgiven H_k\right),
\end{equation}
\normalsize
where every pixel $(ij)$ is modeled through a set of mixture components \mbox{$\mathdutchcal{g}\left(X_{(ij)} \sgiven H_k\right) := p\left(X_{ij} | Z_{k(ij)}=1, H_k\right)$} that model how pixels of the individual sub-images $I_k$ are generated,
as well as through the mixture weights $m_{ij}(H_k):= p\left(Z_{k(ij)}=1| H_k\right)$ that model which point of each object is closest to the camera.

\section{Evidence Lower Bound}  \label{appdx:problem_formulation} \label{appdx:elbo}
Here we provide a derivation of the evidence lower bound. We begin with the log probability of the observations $X^{(1:T)}$ conditioned on a sequence of actions $a^{(0:T-1)}$:
\footnotesize
\begin{align}
    \log p\left(X^{(0:T)} \;\middle|\; a^{(0:T-1)}\right) &= \log \int_{h_{1:K}^{(0:T)}} p\left(X^{(0:T)}, h^{(0:T)}_{1:K} \;\middle|\; a^{(0:T-1)}\right) \; dh_{1:K}^{(0:T)}.\nonumber\\
    &= \log \int_{h_{1:K}^{(0:T)}} p\left(X^{(0:T)}, h^{(0:T)}_{1:K} \;\middle|\; a^{(0:T-1)}\right) \frac{q\left(h^{(0:T)}_{1:K} \given \cdot \right)}{q\left(h^{(0:T)}_{1:K} \given \cdot \right)} \; dh_{1:K}^{(0:T)}.\nonumber\\
    &= \log \mathbb{E}_{h^{(0:T)}_{1:K} \sim q\left(H^{(0:T)}_{1:K} \given \cdot \right)} \left[\frac{p\left(X^{(0:T)}, h^{(0:T)}_{1:K} \;\middle|\; a^{(0:T-1)}\right)}{q\left(h^{(0:T)}_{1:K} \given \cdot \right)}\right]\nonumber\\
    &\geq \mathbb{E}_{h^{(0:T)}_{1:K} \sim q\left(H^{(0:T)}_{1:K} \given \cdot \right)} \log \left[\frac{p\left(X^{(0:T)}, h^{(0:T)}_{1:K} \;\middle|\; a^{(0:T-1)}\right)}{q\left(h^{(0:T)}_{1:K} \given \cdot \right)}\right]. \label{eqn:appdx:elbo}
\end{align}
\normalsize
We have freedom to choose the approximating distribution $q\left(H^{(0:T)}_{1:K} \given \cdot \right)$ so we choose it to be conditioned on the past states and actions, factorized across time:
\footnotesize
\begin{equation*}
    q\left(H^{(0:T)}_{1:K} \given x^{(0:T)}, a^{(0:T)}\right) = q\left(H^{(0)}_{1:K} \sgiven x^{(0)}\right) \prod_{t=1}^T q\left(H^{(t)}_{1:K} \given H^{(t-1)}_{1:K}, x^{(t)}, a^{(t-1)}\right)
\end{equation*}
\normalsize
With this factorization, we can use linearity of expectation to decouple Equation~\ref{eqn:appdx:elbo} across timesteps:
\footnotesize
\begin{equation*}
    \mathbb{E}_{h^{(0:T)}_{1:K} \sim q\left(H^{(0:T)}_{1:K} \sgiven x^{(0:T)}, a^{(0:T)}\right)} \log \left[\frac{p\left(X^{(0:T)}, h^{(0:T)}_{1:K} \;\middle|\; a^{(0:T-1)}\right)}{q\left(h^{(0:T)}_{1:K} \sgiven x^{(0:T)}, a^{(0:T)}\right)}\right] = \sum_{t=0}^{(t)} \mathcal{L}^{(t)}_{\text{r}} - \mathcal{L}^{(t)}_{\text{c}},
\end{equation*}
\normalsize
where at the first timestep
\footnotesize
\begin{align*}
    \mathcal{L}^{(0)}_{\text{r}} &= \mathbb{E}_{h_{1:K}^{(0)} \sim q\left(H_{1:K}^{(0)} \given X^{(0)}\right)}\left[ \log p\left(X^{(0)} \given h_{1:K}^{(0)} \right)\right]\\
     \mathcal{L}^{(0)}_{\text{c}} &= D_{KL}\left(q\left(H_{1:K}^{(0)} \given X^{(0)}\right) \kld p\left(H_{1:K}^{(0)}\right)\right)
\end{align*}
\normalsize
and at subsequent timesteps
\footnotesize
\begin{align*}
    \mathcal{L}^{(t)}_{\text{r}} &= \mathbb{E}_{h^{(t)}_{1:K} \sim q\left(H^{(t)}_{1:K} \sgiven h^{(0:t-1)}_{1:K}, X^{(0:t)}, a^{(0:t-1)} \right)}\left[\log p\left(X^{(t)} \given h^{(t)}_{1:K}\right)\right]\\
    \mathcal{L}^{(t)}_{\text{c}} &= \mathbb{E}_{h^{(t-1)}_{1:K} \sim q\left(H^{(t-1)}_{1:K} \sgiven h^{(0:t-2)}_{1:K}, X^{(1:t-1)}, a^{(0:t-2)} \right)}\left[D_{KL}\left(q\left(H^{(t)}_{1:K} \given h^{(t-1)}_{1:K}, X^{(t)}, a^{(t-1)}\right) \kld p\left(H^{(t)}_{1:K} \given h^{(t-1)}_{1:K}, a^{(t-1)} \right)\right)\right].
\end{align*}
\normalsize
By the Markov property, the marginal $q(H^{(t)}_{1:K} \sgiven h^{(0:t-1)}_{1:K}, X^{(0:t)}, a^{(0:t-1)})$ is computed recursively as
\footnotesize
\begin{equation*}
    \mathbb{E}_{h^{(t-1)} \sim q\left(H^{(t-1)}_{1:K} \sgiven h^{(0:t-2)}_{1:K}, X^{(0:t-1)}, a^{(0:t-2)} \right)}\left[ q\left(H^{(t)}_{1:K} \given h^{(t-1)}_{1:K}, X^{(t)}, a^{(t-1)} \right) \right]
\end{equation*}
\normalsize
whose base case is $q\left(H^{(0)} \,|\, X^{(0)} \right)$ when $t=0$. 

We approximate observation distribution $p(X \sgiven H_{1:K})$ and the dynamics distribution $p(H'_{1:K} \sgiven H_{1:K}, a)$ by learning the parameters of the observation model $\mathdutchcal{G}$ and dynamics model $\mathdutchcal{D}$ respectively as outputs of neural networks.
We approximate the recognition distribution $q(H^{(t)}_{1:K} \sgiven h^{(t-1)}_{1:K}, X^{(t)}, a^{(t-1)})$ via an inference procedure that refines better estimates of the posterior parameters, computed as an output of a neural network.
To compute the expectation in the marginal $q(H^{(t)}_{1:K} \sgiven h^{(0:t-1)}_{1:K}, X^{(0:t)}, a^{(0:t-1)})$, we follow standard practice in amortized variational inference by approximating the expectation with a single sample of the sequence $h^{(0:t-1)}_{1:K}$ by sequentially sampling the latents for one timestep given latents from the previous timestep, and optimizing the ELBO via stochastic gradient ascent~\citep{doersch2016tutorial,kingma2013auto,rezende2014stochastic}.

\section{Posterior Predictive Distribution}
\label{appdx:posterior_predictive_distribution}
Here we provide a derivation of the posterior predictive distribution for the dynamic latent variable model with multiple latent states.
Section~\ref{appdx:elbo} described how we compute the distributions $p(X \sgiven H_{1:K})$, $p(H'_{1:K} \sgiven H_{1:K}, a)$, $q(H^{(t)}_{1:K} \sgiven h^{(t-1)}_{1:K}, X^{(t)}, a^{(t-1)})$, and $q(H_{1:K}^{(0:T)} \sgiven x^{(1:T)}, a^{(1:T)})$.
Here we show that these distributions can be used to approximate the predictive posterior distribution $p(X^{(T+1:T+d)} \sgiven x^{(0:T)}, a^{(0:T+d)})$ by maximizing the following lower bound:
\footnotesize
\begin{align}
    \log p\left(X^{(T+1:T+d)} \given x^{(0:T)}, a^{(0:T+d)}\right) &=\int_{h_{1:K}^{(0:T+d)}} p\left(X^{(T+1:T+d)}, h_{1:K}^{(0:T+d)} \given x^{(0:T)}, a^{(0:T+d)}\right) \;dh_{1:K}^{(0:T+d)}\nonumber\\
    &=\int_{h_{1:K}^{(0:T+d)}} p\left(X^{(T+1:T+d)}, h_{1:K}^{(0:T+d)} \given x^{(0:T)}, a^{(0:T+d)}\right) \frac{q\left(h_{1:K}^{(0:T+d)} \given \cdot \right)}{q\left(h_{1:K}^{(0:T+d)} \given \cdot \right)}\;dh_{1:K}^{(0:T+d)}\nonumber\\
    &=\log \mathbb{E}_{h_{1:K}^{(0:T+d)} \sim q\left(H_{1:K}^{(0:T+d)} \given \cdot \right)} \left[\frac{p\left(X^{(T+1:T+d)}, h_{1:K}^{(0:T+d)} \given x^{(0:T)}, a^{(0:T+d)}\right)}{q\left(h_{1:K}^{(0:T+d)} \given \cdot \right)}\right]\nonumber\\
    &\geq \mathbb{E}_{h_{1:K}^{(0:T+d)} \sim q\left(H_{1:K}^{(0:T+d)} \given \cdot \right)} \log \left[\frac{p\left(X^{(T+1:T+d)}, h_{1:K}^{(0:T+d)} \given x^{(0:T)}, a^{(0:T+d)}\right)}{q\left(h_{1:K}^{(0:T+d)} \given \cdot \right)}\right].\label{eqn:appdx:post_pred_elbo}
\end{align}
\normalsize
The numerator $p(X^{(T+1:T+d)}, h_{1:K}^{(0:T+d)} \sgiven x^{(0:T)}, a^{(0:T+d)})$ can be decomposed into two terms, one of which involving the posterior $p(h_{1:K}^{(0:T+d)} \sgiven x^{(0:T)}, a^{(0:T+d)})$:
\footnotesize
\begin{equation*}
    p\left(X^{(T+1:T+d)}, h_{1:K}^{(0:T+d)} \given x^{(0:T)}, a^{(0:T+d)}\right) = p\left(X^{(T+1:T+d)} \given h_{1:K}^{(0:T+d)}\right) p\left(h_{1:K}^{(0:T+d)} \given x^{(0:T)}, a^{(0:T+d)}\right),
\end{equation*}
\normalsize
This allows Equation~\ref{eqn:appdx:post_pred_elbo} to be broken up into two terms:
\footnotesize
\begin{align*}
    \mathbb{E}_{h_{1:K}^{(0:T+d)} \sim q\left(H_{1:K}^{(0:T+d)} \given \cdot \right)} \log p\left(X^{(T+1:T+d)} \given h_{1:K}^{(0:T+d)}\right) -D_{KL}\left(q\left(H_{1:K}^{(0:T+d)} \given \cdot \right) \kld p\left(H_{1:K}^{(0:T+d)} \given x^{(0:T)}, a^{(0:T+d)}\right) \right)
\end{align*}
\normalsize
Maximizing the second term, the negative KL-divergence between the variational distribution $q(H_{1:K}^{(0:T+d)} \sgiven \cdot)$ and the posterior $p(H_{1:K}^{(0:T+d)} \sgiven x^{(0:T)}, a^{(0:T+d)})$ is the same as maximizing the following lower bound:
\footnotesize
\begin{equation} \label{eqn:appdx:neg_kl}
    \mathbb{E}_{h_{1:K}^{(0:T)} \sim q\left(h_{1:K}^{(0:T)} \given \cdot \right)} \log p\left(x^{(0:T)} \given h_{1:K}^{(0:T)}, a^{(0:T-1)}\right) - D_{KL}\left(q\left(H_{1:K}^{(0:T+d)} \given \cdot \right) \kld p\left(H_{1:K}^{(0:T+d)} \given a^{(0:T+d)}\right) \right)
\end{equation}
\normalsize
where the first term is due to the conditional independence between $X^{(0:T)}$ and the future states $H_{1:K}^{(T+1:T+d)}$ and actions $A^{(T+1:T+d)}$.
We choose to express $q\left(H_{1:K}^{(0:T+d)} \given \cdot \right)$ as conditioned on past states and actions, factorized across time:
\footnotesize
\begin{equation*}
    q\left(H^{(0:T+d)} \given x^{(0:T)}, a^{(0:T+d-1)}\right) = q\left(H^{(0)}_{1:K} \sgiven x^{(0)}\right) \prod_{t=1}^{T+d} q\left(H^{(t)}_{1:K} \given H^{(t-1)}_{1:K}, x^{(t)}, a^{(t-1)}\right).
\end{equation*}
\normalsize
In summary, Equation~\ref{eqn:appdx:post_pred_elbo} can be expressed as
\footnotesize
\begin{align*}
    &\mathbb{E}_{h_{1:K}^{(0:T+d)} \sim q\left(H^{(0:T+d)} \given x^{(0:T)}, a^{(0:T+d-1)}\right)} \log p\left(X^{(T+1:T+d)} \given h_{1:K}^{(0:T+d)}\right)\\
    + &\mathbb{E}_{h_{1:K}^{(0:T)} \sim q\left(H^{(0:T)} \given x^{(0:T)}, a^{(0:T-1)}\right)} \log p\left(x^{(0:T)} \given h_{1:K}^{(0:T)}, a^{(0:T-1)}\right) \\
    - &D_{KL}\left(q\left(H^{(0:T+d)} \given x^{(0:T)}, a^{(0:T+d-1)}\right) \kld p\left(H_{1:K}^{(0:T+d)} \given a^{(0:T+d)}\right) \right)
\end{align*}
\normalsize
which can be interpreted as the standard ELBO objective for timesteps $0:T$, plus an addition reconstruction term for timesteps $T+1:T+d$, a reconstruction term for timesteps $0:T$.
We can maximize this using the same techniques as maximizing Equation~\ref{eqn:appdx:elbo}.

Whereas approximating the ELBO in Equation~\ref{eqn:appdx:post_pred_elbo} can be implemented by rolling out OP3 to predict the next observation via teacher forcing~\citep{williams1989learning}, approximating the posterior predictive distribution in Equation~\ref{eqn:appdx:post_pred_elbo} can be implemented by rolling out the dynamics model $d$ steps beyond the last observation and using the observation model to predict the future observations.

\section{Interactive Inference} \label{appdx:interactive_inference}
Algorithms~\ref{alg:appdx:timestep_0} and~\ref{alg:appdx:timestep_t} detail $M$ steps of the interactive inference algorithm at timestep $0$ and $t \in [1, T]$ respectively.
Algorithm~\ref{alg:appdx:timestep_0} is equivalent to the IODINE algorithm described in~\citep{greff2019iodine}.
Recalling that $\lambda_{1:K}$ are the parameters for the distribution of the random variables $H_{1:K}$, we consider in this paper the case where this distribution is an isotropic Gaussian (e.g. $\mathcal{N}(\lambda_k)$ where $\lambda_k = (\mu_k, \sigma_k)$), although OP3 need not be restricted to the Gaussian distribution.
The \textit{refinement network} $f_{\mathdutchcal{q}}$ produces the parameters for the distribution $\mathdutchcal{q}(H^{(t)}_{k} \sgiven h^{(t-1)}_{k}, x^{(t)}, a^{(t)})$.
The \textit{dynamics network} $f_{\mathdutchcal{d}}$ produces the parameters for the distribution $\mathdutchcal{d}(H^{(t)}_k \sgiven h^{(t-1)}_k, h^{(t-1)}_{[\neq k]}, a^{(t)})$.
To implement $\mathdutchcal{q}$, we repurpose the dynamics model to transform $h_k^{(t-1)}$ into the initial posterior estimate $\lambda_k^{(0)}$ and then use $f_{\mathdutchcal{q}}$ to iteratively update this parameter estimate.
$\beta_k$ indicates the auxiliary inputs into the refinement network used in~\citep{greff2019iodine}.
We mark the major areas where the algorithm at timestep $t$ differs from the algorithm at timestep $0$ in \textcolor{NavyBlue}{blue}.

\begin{algorithm}[H]
    \caption{Interactive Inference: Timestep $0$}
    \label{alg:appdx:timestep_0}
    \footnotesize
    \begin{algorithmic}[1]
        \State \textbf{Input:} observation $x^{(0)}$
        \State \textbf{Initialize:} parameters $\lambda^{(0,0)}$
        \For{$i = 0$ \textbf{to} $M-1$}
            \State Sample $h^{(0,i)}_k \sim \mathcal{N}\left(\lambda^{(0,i)}_k\right)$ for each entity $k$
            \State Evaluate $\mathcal{L}^{(0,i)} \approx \log \mathdutchcal{G}\left(x^{(0)} \,|\, h^{(0,i)}_{1:K} \right) - D_{KL}\left(\mathcal{N}\left(\lambda_{1:K}^{(0,i)}\right) \kld \mathcal{N}\left(0, I\right)\right)$ 
            \State Calculate $\nabla_{\lambda_k} \mathcal{L}^{(0, i)}$ for each entity $k$
            \State Assemble auxiliary inputs $\beta_k$ for each entity $k$
            \State Update $\lambda^{(0, i+1)}_k \gets f_{\text{refine}} \left(x^{(0)}, \nabla_\lambda \mathcal{L}^{(0, i)}, \lambda^{(0, i)}, \beta_k^{(0,i)}\right)$ for each entity $k$
        \EndFor
        \State \Return $\lambda^{(0, M)}$
    \end{algorithmic}
\end{algorithm}

\begin{algorithm}[H]
    \caption{Interactive Inference: Timestep $t$}
    \label{alg:appdx:timestep_t}
    \footnotesize
    \begin{algorithmic}[1] 
        \State \textbf{Input:} observation $x^{(t)}$, \textcolor{NavyBlue}{previous action $a^{(t-1)}$, previous entity states $h^{(t-1)}_{1:K}$}
        \State \textcolor{NavyBlue}{\textbf{Predict} $\lambda^{(t,0)}_k \gets f_{\mathdutchcal{d}}\left(h^{(t-1)}_k, h^{(t-1)}_{[\neq k]}, a^{(t-1)}\right)$} for each entity $k$
        \For{$i = 0$ \textbf{to} $M-1$}
            \State Sample $h_k^{(t,i)} \sim \mathcal{N}\left(\lambda^{(t, i)}\right)$ for each entity $k$
            \State Evaluate $\mathdutchcal{L}^{(t, i)} \approx \log \mathdutchcal{G}\left(x^{(t)} \,|\, h_{1:K}^{(t)} \right) - D_{KL}\left({\mathcal{N}\left(\lambda^{(t,i)}_{1:K}\right) \kld \textcolor{NavyBlue}{\mathcal{N}\left(\lambda^{(t,0)}_{1:K}\right)}}\right)$
            \State Calculate $\nabla_{\lambda_k} \mathdutchcal{L}^{(t, i)}$ for each entity $k$
            \State Assemble auxiliary inputs $\beta_k$ for each entity $k$
            \State Update $\lambda^{(t, i+1)}_k \gets f_{\mathdutchcal{q}}\left(x^{(t)}, \nabla_{\lambda_k} \mathdutchcal{L}^{(t, i)}, \lambda^{(t, i)}_k, \beta_k^{(t,i)}\right)$ for each entity $k$
        \EndFor
        \State \Return $\lambda^{(t, M)}$
    \end{algorithmic}
\end{algorithm}

\paragraph{Training:} We can train the entire OP3 system end-to-end by backpropagating through the entire inference procedure, using the ELBO at every timestep as a training signal for the parameters of $\mathcal{G}$, $\mathcal{D}$, $\mathcal{Q}$ in a similar manner as~\citep{van2018relational}.
However, the interactive inference algorithm can also be naturally be adapted to predict rollouts by using the dynamics model to propagate the $\lambda_{1:K}$ for multiple steps, rather than just the one step for predicting $\lambda_{1:K}^{(t,0)}$ in line 2 of Algorithm~\ref{alg:appdx:timestep_t}.
To train OP3 to rollout the dynamics model for longer timescales, we use a curriculum that increases the prediction horizon throughout training.

\section{Cost Function}
Let $\hat{I}(H_{k}) := m(H_k) \cdot \mathdutchcal{g}\left(X \sgiven H_k\right)$ be a \textit{masked} sub-image (see Appdx:~\ref{appdx:obs_model}).
We decompose the cost of a particular configuration of objects into a distance function between entity states, $\mathdutchcal{c}(H_a, H_b)$. For the first environment with single-step planning we use $L_2$ distance of the corresponding masked subimages: $\mathdutchcal{c}(H_a, H_b) = L_2(\hat{I}(H_a), \hat{I}(H_b))$.
For the second environment with multi-step planning we a different distance function since the previous one may care more about if a shape matches than if the color matches. We instead use a form of intersection over union but that counts intersection if the mask aligns and pixel color values are close $\mathdutchcal{c}(H_a, H_b) = 1 -  \frac{{\sum_{i,j} m_{ij}(H_a) > 0.01 \text{ and } m_{ij}(H_b) > 0.01} \text{ and } L_2(\mathdutchcal{g}(H_a)_{(ij)}, \mathdutchcal{g}(H_b)_{(ij)}) < 0.1}{\sum_{i,j} m_{ij}(H_a) > 0.01 \text{ or } m_{ij}(H_b) > 0.01}$. We found this version to work better since it will not give low cost to moving a wrong color block to the position of a different color goal block.

\section{Architecture and Hyperparameter Details} \label{appdx:arch_details}
We use similar model architectures as in \citep{greff2019iodine} and so have rewritten some details from their appendix here. Differences include the dynamics model, inclusion of actions, and training procedure over sequences of data.  
Like~\citep{hafner2018learning}, we define our latent distribution of size $R$ to be divided into a deterministic component of size $R_d$ and stochastic component of size $R_s$.
We found that splitting the latent state into a deterministic and stochastic component (as opposed to having a fully stocahstic representation) was helpful for convergence.
We parameterize the distribution of each $H_k$ as a diagonal Gaussian, so the output of the refinement and dynamics networks are the parameteres of a diagonal Gaussian.
We parameterize the output of the observation model also as a diagonal Gaussian with means $\mu$ and global scale $\sigma=0.1$. The observation network outputs the $\mu$ and mask $m_k$.

\textbf{Training:} All models are trained with the ADAM optimizer \citep{kingma2014adam} with default parameters and a learning rate of 0.0003. We use gradient clipping as in \citep{Pascanu2012UnderstandingTE} where if the norm of global gradient exceeds 5.0 then the gradient is scaled down to that norm. 

\textbf{Inputs:} 
For all models, we use the following inputs to the refinement network, where LN means Layernorm and SG means stop gradients.
The following image-sized inputs are concatenated and fed to the corresponding convolutional network: 

\begin{center}
\begin{tabular}{llccc}
\toprule
    Description & Formula & LN & SG & Ch.\\ \midrule
    image & $X$ & & & 3 \\
    means & $\mu$ &  & & 3  \\
    mask  & $m_k$ & & & 1 \\
    mask-logits & $\hat{m}_k$ & & & 1 \\
    mask posterior & $p(m_k|X,\mu)$ & & & 1 \\
    gradient of means & $\nabla_{μ_k}\mathcal{L}$ & \checkmark & \checkmark & 3 \\
    gradient of mask & $\nabla_{m_k}\mathcal{L}$ & \checkmark & \checkmark & 1 \\
    pixelwise likelihood & $p(X \sgiven H)$ & \checkmark & \checkmark & 1 \\
    leave-one-out likelih. & $p(X \sgiven H_{i \neq k})$ & \checkmark & \checkmark & 1 \\
    coordinate channels & & & & 2 \\
    \midrule
    \multicolumn{4}{r}{total:} & 17 \\
    \bottomrule
\end{tabular}
\end{center}

\subsection{Observation and Refinement Networks}
The posterior parameters $\lambda_{1:K}$ and their gradients are flat vectors, and we concatenate them with the output of the convolutional part of the refinement network and use the result as input to the refinement LSTM:

\begin{center}
\begin{tabular}{llcc}
\toprule
    Description & Formula & LN & SG\\ \midrule
    gradient of posterior & $\nabla_{\lambda_k}\mathcal{L}$ & \checkmark & \checkmark \\
    posterior & $\lambda_k$ & &\\
    \bottomrule
\end{tabular}
\end{center}

All models use the ELU activation function and the convolutional layers use a stride equal to 1 and padding equal to 2 unless otherwise noted. For the table below $R_s=64$ and $R=128$.

\begin{center}
\begin{tabular}{lccl}
    \multicolumn{4}{c}{\textbf{Observation Model Decoder}}\\ 
    \toprule
    Type & Size/\textit{Ch.} & Act. Func. & Comment\\ \midrule
    Input: $H_i$  & $R$ &  & \\
    Broadcast & \textit{$R$+2} & & + coordinates\\
    Conv $5\times5$ & \textit{32} & ELU & \\
    Conv $5\times5$ & \textit{32} & ELU & \\
    Conv $5\times5$ & \textit{32} & ELU & \\
    Conv $5\times5$ & \textit{32} & ELU & \\
    Conv $5\times5$ & \textit{4} & Linear  & RGB + Mask\\
    \bottomrule
\end{tabular}
\end{center}

\begin{center}
\begin{tabular}{lccl}
    \multicolumn{4}{c}{\textbf{Refinement Network}}\\
    \toprule
    Type & Size/\textit{Ch.} & Act. Func. & Comment\\ \midrule
    MLP & 128 & Linear & \\
    LSTM & 128 & Tanh & \\
    Concat $[{\lambda_i}, \nabla_{\lambda_i}]$ & $2R_s$ & & \\
    MLP & 128 & ELU & \\
    Avg. Pool & $R_s$ & \\
    Conv $3\times3$ & \textit{$R_s$}  & ELU & \\
    Conv $3\times3$ & \textit{32}  & ELU & \\
    Conv $3\times3$ & \textit{32}  & ELU & \\
    Inputs & \textit{17}  & & \\
    \bottomrule
\end{tabular}
\end{center}

\subsection{Dynamics Model} \label{appdx:dyn_model}
The dynamics model $\mathcal{D}$ models how each entity $H_k$ is affected by action $A$ and the other entity $H_{[\neq k]}$.
It applies the same function $\mathdutchcal{d}(H_k' \sgiven H_k, H_{[\neq k]}, A)$ to each state, composed of several functions illustrated and described in Fig.~\ref{fig:dynamics}:
\begin{footnotesize}
\begin{align*}
&\tilde{H}_k = d_{o}(H_k) \qquad
\tilde{A} = d_{a}(A^t) \qquad
\tilde{H}_k^{\text{act}} = d_{ao}(\tilde{H}_k \tilde{A}) \\
&H_k^{\text{interact}} = \sum_{i \neq k}^K d_{oo}(\tilde{H_i}^{\text{act}}, \tilde{H_k}^{\text{act}}) \qquad
H_{k}^{'} = d_{\text{comb}}(\tilde{H}_k^{\text{act}}, H_k^{\text{interact}}),
\end{align*}
\end{footnotesize}
where for a given entity $k$, $d_{ao}(\tilde{H}_k \tilde{A}) := d_{\text{act-eff}}(\tilde{H}_k, \tilde{A}) \cdot d_{\text{act-att}}(\tilde{H}_k, \tilde{A})$ computes how $(d_{\text{act-eff}})$ and to what degree $(d_{\text{act-att}})$ an action affects the entity and $d_{oo}(\tilde{H_i}^{\text{act}}, \tilde{H_k}^{\text{act}}) := d_{\text{obj-eff}}(\tilde{H}_i^{\text{act}}, \tilde{H}_k^{\text{act}}) \cdot d_{\text{obj-att}}(\tilde{H}_i^{\text{act}}, \tilde{H}_k^{\text{act}}
)$ computes how $(f_{\text{obj-eff}})$ and to what degree $(d_{\text{obj-att}})$ other entities affect that entity.
$d_{\text{obj-eff}}$ and $d_{\text{obj-att}}$ are shared across all entity pairs. The other functions are shared across all entities. The dynamics network takes in a sampled state and outputs the parameters of the posterior distribution. Similar to~\citep{hafner2018learning} the output $H'_k$ is then split into deterministic and stochastic components each of size 64 with separate networks $f_{\text{det}}$ and $f_{\text{sto}}$. All functions are parametrized by single layer MLPs.

\begin{center}
\begin{tabular}{lccl}
    \multicolumn{4}{c}{\textbf{Dynamics Network}}\\
    \toprule
    Function  & Output & Act. Func. & MLP Size\\ \midrule
    $d_o(H_k$) & $\tilde{H_k}$ & ELU & 128 \\
    $d_a(A)$ & $\tilde{A}$ & ELU &  32 \\
    $d_{\text{act-eff}}(\tilde{H_k}, \tilde{A})$ &  & ELU & 128 \\
    $d_{\text{act-att}}(\tilde{H_k}, \tilde{A})$ &  & Sigmoid & 128 \\
    $d_{\text{obj-eff}}(\tilde{H}_i^{\text{act}}, \tilde{H}_j^{\text{act}})$ &  & ELU & 256 \\
    $d_{\text{obj-att}}(\tilde{H}_i^{\text{act}}, \tilde{H}_k^{\text{act}})$ &  & Sigmoid & 256 \\
    $d_{\text{comb}}(\tilde{H}_i^{\text{act}}, \tilde{H}_k^{\text{interact}})$ &  $H'_k$& ELU & 256 \\
    $f_{\text{det}}(H'_k)$ &  $H'_{\text{k,det}}$&  & 128 \\
    $f_{\text{sto}}(H'_k)$ &  $H'_{\text{k,sto}}$&  & 128 \\
    \bottomrule
\end{tabular}
\end{center}

This architectural choice for the dynamics model is an action-conditioned modification of the interaction function used in Relational Neural Expectation Maximization (RNEM)~\citep{van2018relational}, which is a latent-space attention-based modification of the Neural Physics Engine (NPE)~\citep{chang2016compositional}, which is one of a broader class of architectures known as graph networks~\citep{battaglia2018relational}.

\section{Experiment Details}

\subsection{Single-Step Block-Stacking}
The training dataset has 60,000 trajectories each containing before and after images of size 64x64 from \cite{janner2018reasoning}. Before images are constructed with actions which consist of choosing a shape (cube, rectangle, pyramid), color, and an $(x, y, z)$ position and orientation for the block to be dropped. At each time step, a block is dropped and the simulation runs until the block settles into a stable position. The model takes in an image containing the block to be dropped and must predict the steady-state effect. Models were trained on scenes with 1 to 5 blocks with $K = 7$ entity variables. The cross entorpy method (CEM) begins from a uniform distribution on the first iteration, uses a population size of 1000 samples per iteration, and uses 10\% of the best samples to fit a Gaussian distribution for each successive iteration.

\begin{figure}[h!]
    \centering
    \includegraphics[width=1\columnwidth]{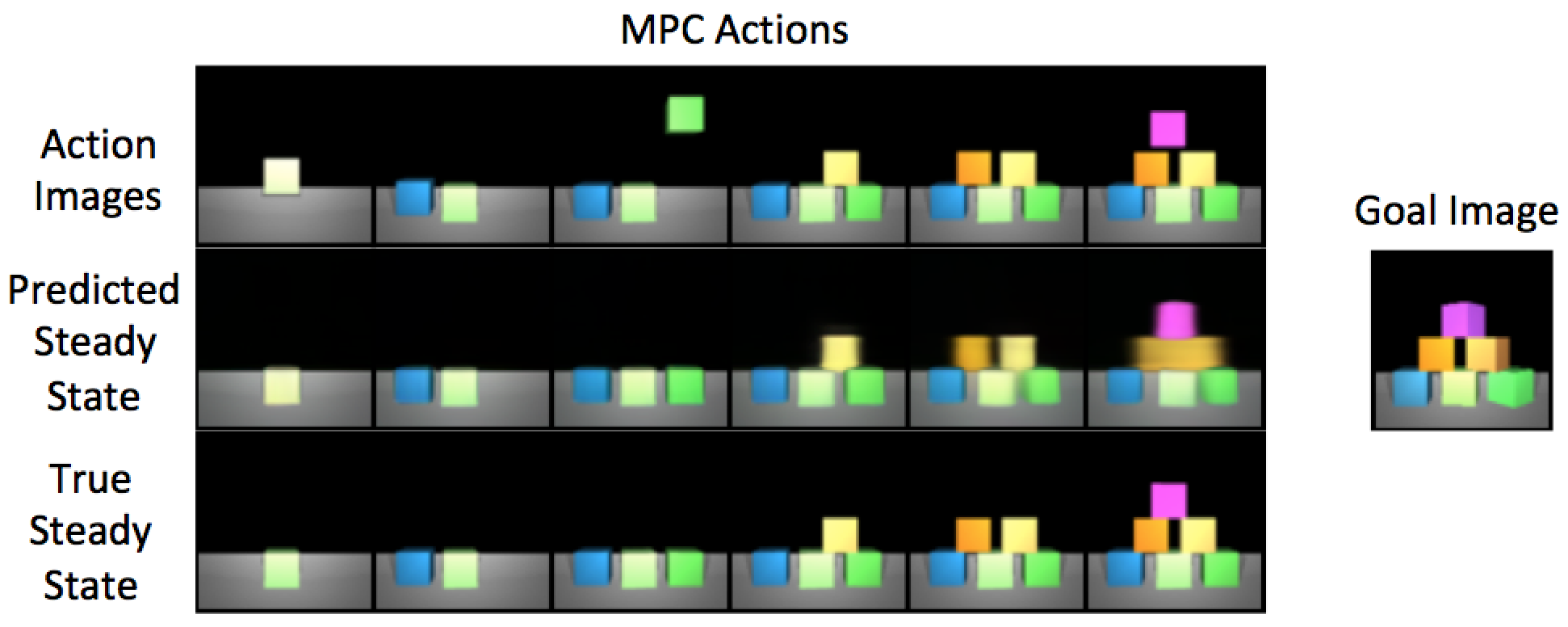}
    \caption{Qualitative results on building a structure from the dataset in~\citep{janner2018reasoning}. The input is an "action image," which depicts how an action intervenes on the state by raising a block in the air. OP3 is trained to predict the steady-state outcome of dropping the block. We see how OP3 is able to accurately and consistently predict the steady state effect, successively capturing the effect of inertial dynamics (gravity) and interactions with other objects.}   \label{fig:mpc_results}
\end{figure}

\begin{figure}[t]
    \centering
    \hspace{-4pt}
    \includegraphics[width=0.5\columnwidth]{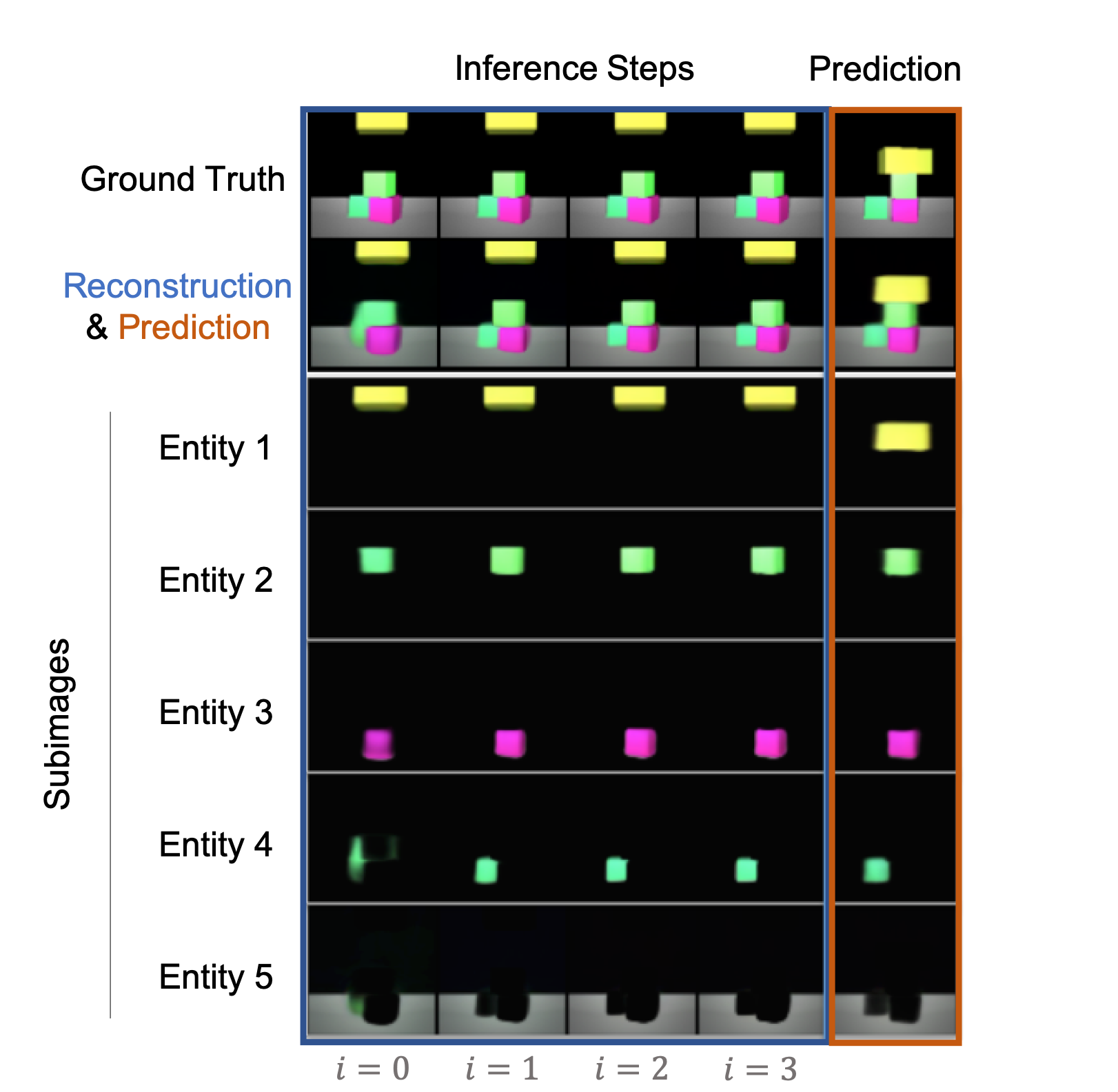}
    \caption{We show a demonstration of a rollout for the dataset from~\citep{janner2018reasoning}. The first four columns show inference iterations (refinement steps) on the single input image, while the last column shows the predicted results using the dynamics module on the learnt hidden states. The bottom 5 rows show the subimages of each entity at each iteration, demonstrating how the model is able to capture individual objects, and the dynamics afterwards. Notice that OP3 only predicts a change in the yellow block while leaving the other latents unaffected. This is a desriable property for dynamics models that operate on scenes with multiple objects.}
    \label{fig:predicted_components}
\end{figure}

\subsection{Multi-Step Block-Stacking} \label{appdx:multi_step_block_stacking}
The training dataset has 10,000 trajectories each from a separate environment with two different colored blocks. Each trajectory contains five frames (64x64) of randomly picking and placing blocks. We bias the dataset such that 30\% of actions will pick up a block and place it somewhere randomly, 40\% of actions will pick up a block and place it on top of a another random block, and 30\% of actions contain random pick and place locations. Models were trained with K = 4 slots. We optimize actions using CEM but we optimize over multiple consecutive actions into the future executing the sequence with lowest cost. For a goal with $n$ blocks we plan $n$ steps into the future, executing $n$ actions. We repeat this procedure $2n$ times or until the structure is complete. Accuracy is computed as $\frac{\text{\# blocks in correct position}}{\text{\# goal blocks}}$, where a correct position is based on a threshold of the distance error. 
\begin{figure}[h!]
    \centering
    \includegraphics[width=0.8\columnwidth]{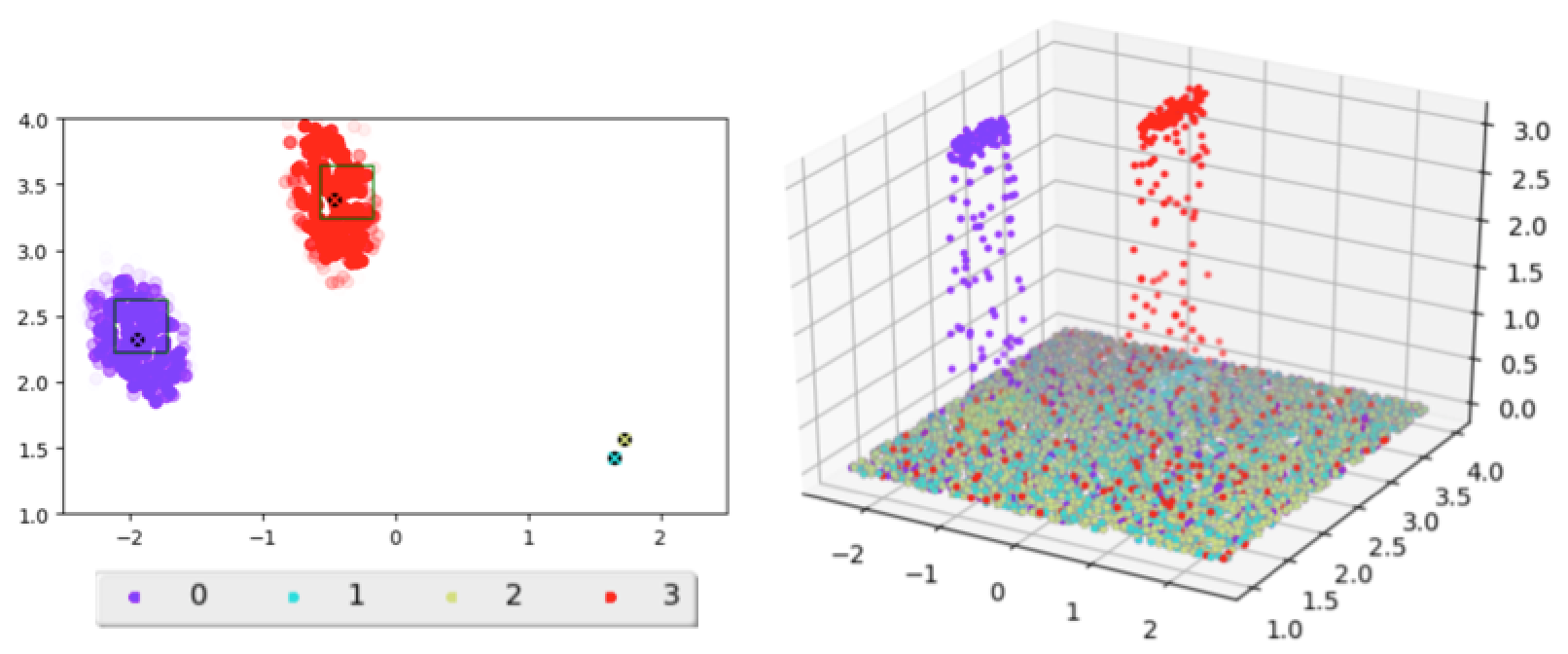}
    \caption{Two-dimensional (left) and three-dimensional (right) visualization of attention values where colors correspond to different latents. The blocks are shown as the green squares in the 2D visualizatio; picking anywhere within the square automatically picks the block up. The black dots with color crosses denote the computed $\texttt{pick\_xy}$ for a given $h_k$. We see that although the individual values are noisy, the means provide good estimates of valid pick locations. In the right plot we see that attention values for all objects are mostly 0, except in the locations corresponding to the objects (purple and red).}   \label{fig:attention_vizualization}
\end{figure}

For MPC we use two difference action spaces:

\textbf{Coordinate Pick Place:} The normal action space involves choosing a pick (x,y) and place (x,y) location. 

\textbf{Entity Pick Place:} A concern with the normal action space is that successful pick locations are sparse (~2\%) given the current block size. Therefore, the probability of picking $n$ blocks consecutively becomes $0.02^n$ which becomes improbable very fast if we just sample pick locations uniformly. We address this by using the pointers to the entity variables to create an action space that involves directly choosing one of the latent entities to move and then a place $(x,y)$ location. This allows us to easily pick blocks consecutively if we can successfully map a latent \texttt{entity\_id} of a block to a corresponding successful pick location. In order to determine the pick $(x,y)$ from an \texttt{entity\_id} $k$, we sample coordinates uniformly over the pick $(x,y)$ space and then average these coordinates weighted by their attention coefficient on that latent:
$$
\texttt{pick\_xy} | h_k = \frac{\sum_{x',y'} p(h_k|x,y) * \texttt{pick\_x'y'} } {\sum_{x',y'} p(h_k|x',y')}
$$
where $p(h_k|x,y)$ are given by the attention coefficients produced by the dynamics model given $h_k$ and the pick location $(x,y)$ and $x',y'$ are sampled from a uniform distribution.
The attention coefficient of $H_k$ is computed as $\sum_{i\neq k}^K d_{\text{obj-att}}(\tilde{H}_i^{\text{act}}, \tilde{H}_k^{\text{act}})$ (see Appdx.~\ref{appdx:dyn_model})

\section{Ablations}
We perform ablations on the block stacking task from \cite{janner2018reasoning} examining components of our model. Table~\ref{tab:stage1_ablations} shows the effect of non-symmetrical models or cost functions. The ``Unfactorized Model'' and ``No Weight Sharing'' follow (c) and (d) from Figure~\ref{fig:combined_overview} and are unable to sufficiently generalize.
``Unfactorized Cost'' refers to simply taking the mean-squared error of the compositie prediction image and the goal image, rather than decomposing the cost per entity masked subimage.
We see that with the same OP3 model trained on the same data, not using an entity-centric factorization of the cost significantly underperforms a cost function that does decompose the cost per entity (c.f. Table~\ref{tab:mse_results}).
\begin{table}[h]
    \centering
    \begin{tabular}{ccc}
    \toprule
         No Weight Sharing & Unfactorized Model & Unfactorized Cost \\
     \midrule
          0 \% & 0 \% & 5\%\\
    \bottomrule
    \end{tabular}
    \captionof{table}{\small{Accuracy of ablations. The no weight sharing model did not converge during training.}}
    \label{tab:stage1_ablations}
\end{table}

\section{Interpretability}
We do not explicitly explore interpretability in this work, but we see that an entity-factorized model readily lends itself to be interpretable by construction. The ability to decompose a scene into specific latents, view latents invididually, and explicitly see how these latents interact with each other could lead to significantly more interpretable models than current unfactorized models. Our use of attention values to determine the pick locations of blocks scratches the surface of this potential. Additionally, the ability to construct cost functions based off individual latents allows for more interpretable and customizable cost functions.
\end{document}